\documentclass[fleqn,10pt]{wlscirep}
\usepackage{graphicx} 
\usepackage{amsmath} 
\usepackage{amssymb}
\usepackage{mathtools}
\usepackage[export]{adjustbox} 

\usepackage{xcolor}
\usepackage[dvipsnames]{xcolor}
\usepackage{cleveref}

\usepackage{lineno}

\newcommand{\net}{\textit{Spotted}}

\usepackage{subcaption}
\crefname{figure}{fig.}{figs.}
\crefname{equation}{equation}{equations}
\Crefname{figure}{Fig.}{Figs.}
\crefname{table}{Table}{Tables}
\Crefname{table}{Table}{Tables}

\renewcommand{\eqref}[1]{Eq.~(\ref{#1})}
\newcommand{\figref}[1]{Fig.~\ref{#1}}

\title{Spotted: Location-informed Reidentification of Hyenas and Leopards in Camera Trap Surveys}

\author[1,*]{Halil Sina Kelebek}
\author[1,2]{Julia Hindel}
\author[5]{Kobus Hoffman}
\author[5]{Lauren Hoffman}
\author[3]{Andrew Loveridge}
\author[6]{Bob Mandinyenya}
\author[5]{Kudakwashe Ncube}
\author[3]{Justin Seymour-Smith}
\author[3]{Andrea Sibanda}
\author[2]{Abhinav Valada}
\author[3]{Matthew Wijers}
\author[1,4]{Daniele De Martini}

\affil[1]{Oxford Robotics Institute, University of Oxford, Oxford, UK}
\affil[2]{Department of Computer Science, University of Freiburg, Germany}
\affil[3]{Wildlife Conservation Research Unit, Department of Biology, University of Oxford, Oxford, UK}
\affil[4]{Oxford e-Research Centre, University of Oxford, Oxford, UK}
\affil[5]{Bubye Valley Conservancy, Zimbabwe}
\affil[6]{Gonarezhou National Park, Chiredzi, Zimbabwe}
\affil[*]{email: halil@robots.ox.ac.uk}

\begin{abstract}

Animal re-identification (ReID) in camera-trap surveys remains challenging due to low image quality, strong variation in illumination and viewpoint, and highly imbalanced numbers of observations per individual. As a result, current ReID performance is often insufficient for fully automated use, and practical workflows typically depend on expert review of algorithmically proposed candidate matches.
Moreover, most existing approaches focus almost exclusively on visual cues and overlook auxiliary information routinely available in field studies, such as image timestamps and camera-trap locations. We introduce \net, a location-informed, human-in-the-loop animal ReID framework that integrates visual similarity with spatio-temporal feasibility priors derived from camera locations, thereby reducing the amount of required expert review. Our method (i) computes an image-model-agnostic feasibility score based on the minimum travel speed required for two detections to correspond to the same individual, (ii) uses these feasibility cues as pseudo-supervision to train a lightweight head on top of a frozen visual foundation model, and (iii) fuses adapted visual similarity with spatio-temporal feasibility to obtain a robust pairwise matching score. We additionally integrate an active pair sampling strategy to accelerate annotation by initially prioritizing uncertain predictions. We evaluate \net{} on three challenging camera-trap ReID datasets comprised of spotted hyenas and leopards, which we release as part of this work. Our model improves average top-5 identification accuracy by 9pp, 2pp and 9pp over the best baseline on our LeopardID102, SpottedHyenaID109 and SpottedHyenaID415 datasets, respectively. Further, we show that our human-in-the-loop strategy reduces the number of queried comparisons by up to 69pp while achieving equivalent positive matches.

\end{abstract}

\begin{document}
\maketitle
\section*{Introduction}

Camera traps are a central tool in wildlife ecology, enabling non-invasive monitoring of animal populations in the wild~\cite{10.3389/fevo.2021.617996}. A challenge in camera-trap-based studies is associating repeated sightings of the same individual and differentiating between separate individuals within a given survey. 
This task, known as animal re-identification (ReID), remains a challenging computer vision problem, and current automated methods still produce unreliable predictions for many species and survey conditions.
Trap-camera imagery is often noisy and exhibits high variation in illumination, animal scale, and viewpoint. Additionally, real-world wildlife monitoring datasets are typically highly imbalanced, with some individuals observed frequently while others are captured only once during the survey period~\cite{adam2024seaturtleid2022,adam2025wildlifereid}. This imbalance results in limited and biased training data, which drastically hinders the performance of purely vision-based models. However, camera-trap surveys possess a structural property: images are captured with a fixed network of camera traps with known spatial locations and timestamps, providing a rich supervisory signal that current state-of-the-art animal reID networks overlook. This metadata encodes implicit constraints on the physical feasibility of the imaged identity. For example, two images captured at distant camera traps within a short time interval imply an implausible travel speed and, consequently, are unlikely to be identity-consistent detections. More importantly, repeated captures at spatially close camera traps over extended time periods reflect stable movement patterns of individual animals. Our ecological motivation for spatial priors is further grounded in the concept of home range~\cite{10.1644/11-MAMM-S-177.1}: individuals of many species tend to remain within spatially bounded regions over extended periods, making spatial proximity an informative prior for identity. Consequently, these limitations motivate a tailored solution that leverages rich spatial priors beyond a simple binary feasibility check.

Wildlife ecology demands flawless animal re-identification to reliably extract meaningful animal movement patterns. Fully automatic identification typically cannot meet this requirement and current workflows rely on human verification of algorithm-proposed image matches for the same or different individual. To avoid an exhaustive annotation effort over a large set of candidate matches~\cite{10.3389/fevo.2021.617996}, we motivate an active pair sampling strategy that initially prioritises uncertain candidate matches for human verification, reducing the total annotation burden while maximising the discovery of true identity matches. At an annotation rate of 5.78 queries per minute, as measured from an expert annotator in our study, the choice of pair selection strategy has a direct and measurable impact on total labeling time. 

In this work, we propose \net, a location-informed, human-in-the-loop animal re-identification framework that integrates visual similarity with spatio-temporal feasibility priors derived from the camera location and timestamp metadata. Rather than treating location priors only as a hard constraint or post-hoc filter, we incorporate them as a continuous signal that shapes both representation learning and pairwise similarity scoring. Our approach consists of three components. First, we introduce a vision-agnostic, pairwise speed-based feasibility scoring that encodes the minimum plausible speed for two detections to correspond to the same individual. Second, we train a lightweight MLP head on top a frozen animal ReID backbone using location-derived pseudo-labels to suppress physically implausible image matches without requiring identity supervision. Third, we fuse the adapted visual similarities with the spatio-temporal feasibility scores and integrate an active pair-sampling strategy to improve retrieval accuracy while considerably reducing the number of pairwise comparisons required in a human-in-the-loop workflow. To evaluate our work, we release three camera-trap ReID datasets (LeopardID102, SpottedHyenaID109, and SpottedHyenaID415) with this work, comprising over $3,200$ camera images taken of $620$ individual animals at $230$ locations collected over a 2-month survey period. Our proposed method demonstrates consistent improvements over SOTA animal re-identification baselines.

\section*{Related Work}

\subsection*{Animal Re-Identification}

Seminal works in animal ReID such as HotSpotter \cite{crall2013hotspotter} rely on handcrafted local feature descriptors to perform instance recognition by matching sparse SIFT-based descriptors~\cite{Lowe2004} and ranking candidate matches based on geometric consistency. Despite its advantages in low data requirements and interpretability, this approach exhibits poor robustness to severe appearance and viewpoint variations. With the successes of Deep Learning techniques over the past decade, the field has shifted towards metric learning approaches that directly map images into an embedding space where intra-identity distances are minimized and inter-identity distances are maximized, resulting in embedding vectors corresponding to the same individual being mapped in proximity~\cite{GUO2020101412,yu2024addressing,hou2025openanimals}. Although many previous works represent individuals with a single global feature vector extracted by passing an image in full through a backbone network~\cite{miewid, Cermak_2024_WACV}, selected works apply local, part-aware approaches to improve robustness across various viewpoints by focusing on local regions or salient parts~\cite{zhu2022pass,yu2024addressing,moskvyak2021keypoint,moskvyak2020learning}. 
The hybrid method proposed by Shukla et al.~\cite{Shukla_2019_ICCV} applies a finetuned DenseNet~\cite{huang2017densely} to produce an initial set of pairwise matches, followed by a SIFT-based re-ranking approach to combat poor image quality and limited training data. 

Another trend is the emergence of multispecies pretraining and foundation models for animal ReID, which aim to produce transferable embeddings across species and datasets. MegaDescriptor~\cite{Cermak_2024_WACV} trains a SWIN backbone~\cite{swin} with an ArcFace loss~\cite{arcface} for animal ReID, providing a competitive baseline. Similarly, the multispecies identification model, MiewID~\cite{miewid} is trained with the ArcFace loss but rather uses the Convolutional Neural Network (CNN) based EfficientNetV2~\cite{pmlr-v139-tan21a} architecture as their model backbone and makes use of a larger, community-curated dataset for model training.

\subsection*{Spatio-Temporal Auxiliary Supervision}
The importance of meta-data for Animal ReID is increasingly explored. A recent animal ReID dataset introduces meta-features such as temperature, circadian rhythm, and orientation~\cite{10.1145/3746027.3758249}. Alongside the dataset, the authors propose the Meta-Feature Adapter~\cite{10.1145/3746027.3758249}, which incorporates metadata into their ReID model as textual input via image and text encoders. Camera traps provide a fixed geometry for the sensor network with known relative positions and timestamps, imposing strong constraints on whether two detections of the same individual are physically plausible. A previous camera trap study with dense camera placement show that timestamps and camera identities can be used directly to reconstruct animal movement paths across a survey area~\cite{Kays2021}, demonstrating that the spatial sensor structure contains rich information about individual movement. Spatio-temporal auxiliary data have been previously explored in tangential domains like fine grained classification problems such as species classification~\cite{mac2019presence} A previous work has explored spatio-temporal priors in a closed set animal ReID setting~\cite{foreground-background}, however requiring pre-labelled class-level training data for each individual due to the closed-set ReID setting, making it unsuitable for the task of unlabelled dataset annotation. While temporal meta-data is often available~\cite{adam2025wildlifereid}, many publicly available animal ReID datasets do not include fine-grained geo-spatial coordinates ~\cite{adam2025wildlifereid,rosa2025gcn,Li2019AmurTR,bpct2022leopard}.

\subsection*{Active Pair Sampling}
Fully automated animal ReID does not achieve flawless accuracy and is therefore often combined with human verification of algorithmically proposed candidate matches~\cite{schneider2019past,Osner_TrapTagger}. This workflow motivates approaches that minimize human annotation needs while optimizing correctly identified individuals.
Active pair sampling has been explored in animal ReID~\cite{sani2025activelearninganimalreidentification} and species identification~\cite{active-learning-camera-trap} as well as person ReID~\cite{jin2022fewerlabelssupportpair,liu2019deep}, however, with a focus on reducing the training set required for model training rather than to speed up complete dataset annotation~\cite{jin2022fewerlabelssupportpair,liu2019deep,active-learning-camera-trap}. Many active learning methods propose clustering approaches to discover informative samples for labeling, with specific strategies varying in how cluster structure is exploited: previous works have explored sampling strategies focusing on on cluster boundaries~\cite{jin2022fewerlabelssupportpair}, distance to core points~\cite{active-cluster,cons-dbscan}, or cluster mismatch between different clustering algorithms~\cite{sani2025activelearninganimalreidentification}. SPAL~\cite{jin2022fewerlabelssupportpair} proposes an uncertainty-driven strategy that clusters samples and queries pairs near cluster boundaries to efficiently improve representation quality with limited labels. DRAL~\cite{liu2019deep} proposes a reinforcement learning based approach to propose uncertain pairs for labeling. Continual learning approaches have also been proposed to continuously update animal species classification models~\cite{Miao2021} and animal ReID models~\cite{Bodesheim2022,sani2025activelearninganimalreidentification} using expert annotations for selected images, thereby enabling incremental performance improvements in model accuracy as new data is collected. Active selection strategies have also been studied more broadly in the animal ReID domain for population size estimation under limited supervision~\cite{counting-clusters}. Across these works, however, the common objective is to reduce the number of labels needed to train or estimate a model or population statistic, rather than to minimise the annotation effort required to fully label a dataset. The problem of accelerating complete dataset annotation by reducing redundant pairwise comparisons for a human expert remains comparatively under explored.

\section*{Data}
\begin{figure*}[t!]
    \centering
    
    \begin{subfigure}[t]{0.33\textwidth}
        \centering
        \clipbox{0cm 0pt 0pt 0.38cm}{
        \includegraphics[width=\dimexpr\linewidth-3pt]{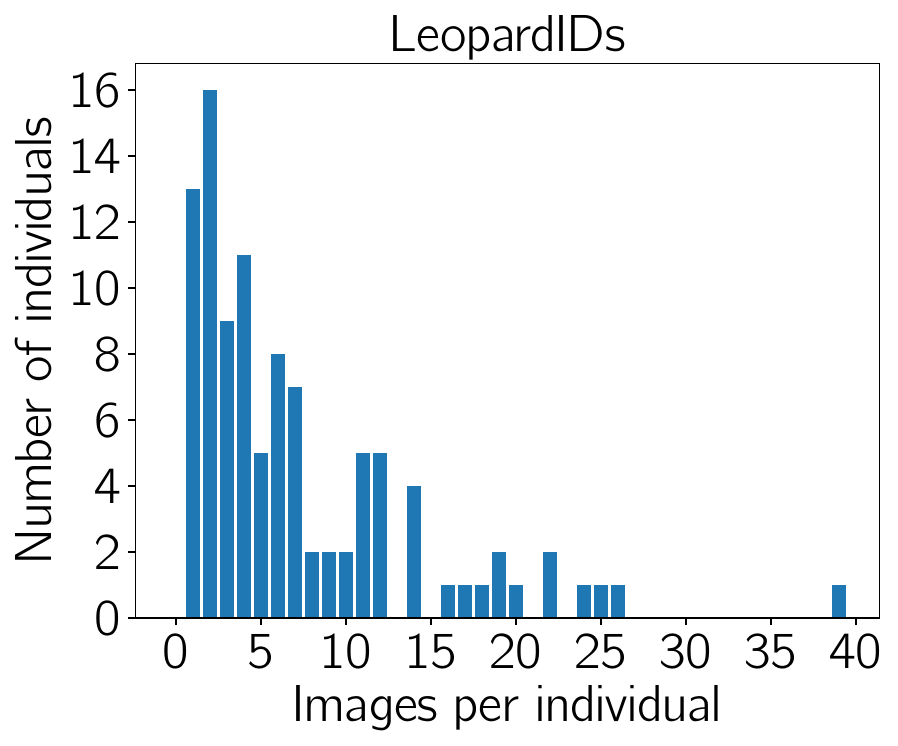}}
        \caption{LeopardID102}
    \end{subfigure}
    \begin{subfigure}[t]{0.33\textwidth}
        \centering
        \clipbox{0.5cm 0pt 0pt 0.38cm}{
        \includegraphics[width=\dimexpr\linewidth-3pt]{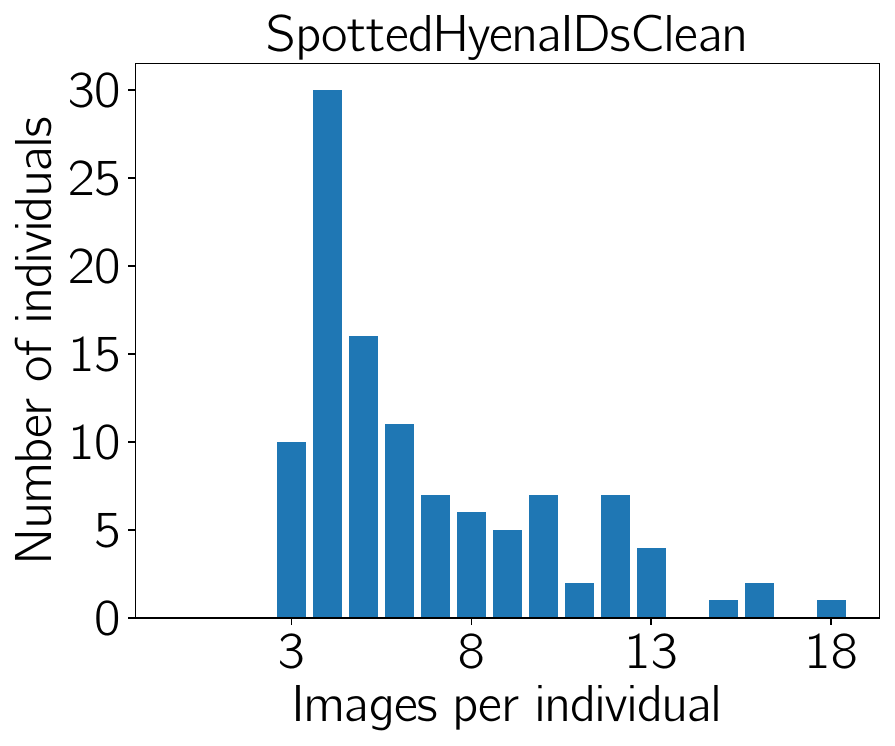}}
        \caption{SpottedHyenaID109}
    \end{subfigure}
    \begin{subfigure}[t]{0.33\textwidth}
        \centering
        \clipbox{0.5cm 0pt 0pt 0.38cm}{
        \includegraphics[width=\dimexpr\linewidth-3pt]{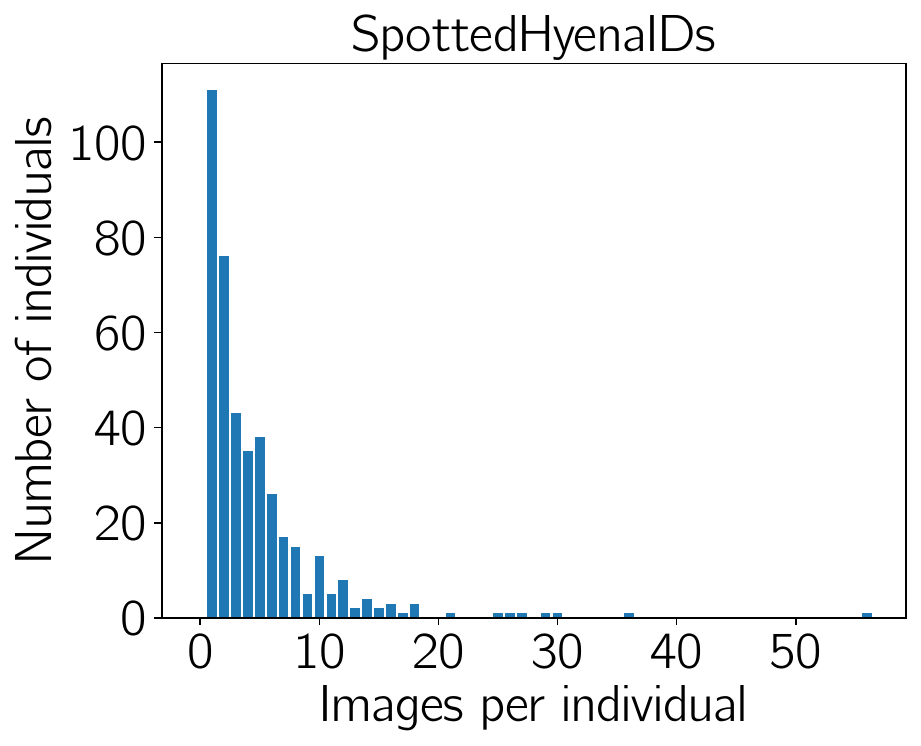}}
        \caption{SpottedHyenaID415}
    \end{subfigure}%

    \begin{subfigure}[t]{\textwidth}
        \centering
        \includegraphics[trim={0 0 0 0cm},clip,width=\textwidth]{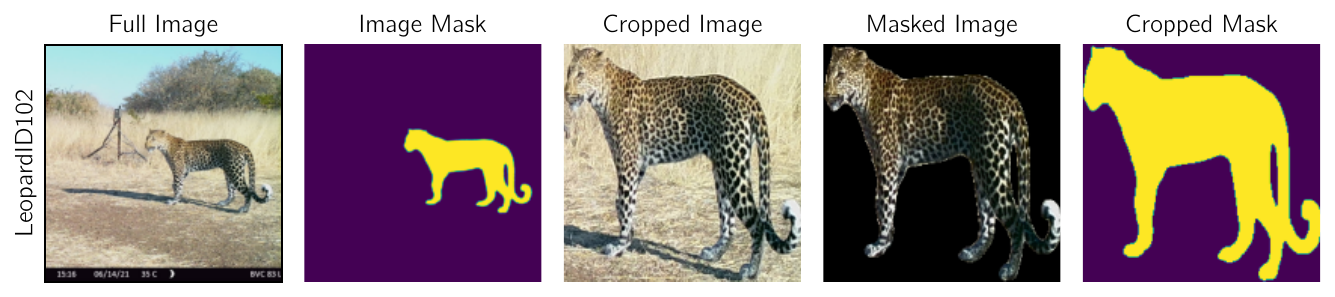} \\[2pt] \includegraphics[trim={0 0 0 0.8cm},clip,width=\textwidth]{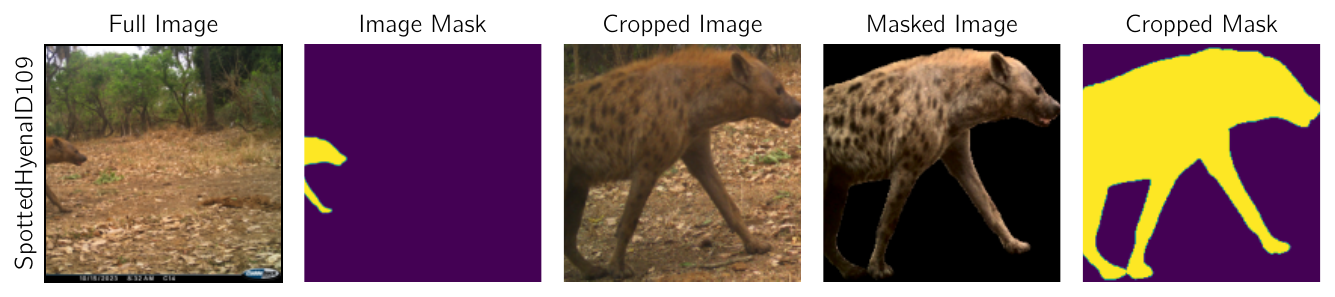} \\[2pt]
        \includegraphics[trim={0 0 0 0.8cm},clip,width=\textwidth]{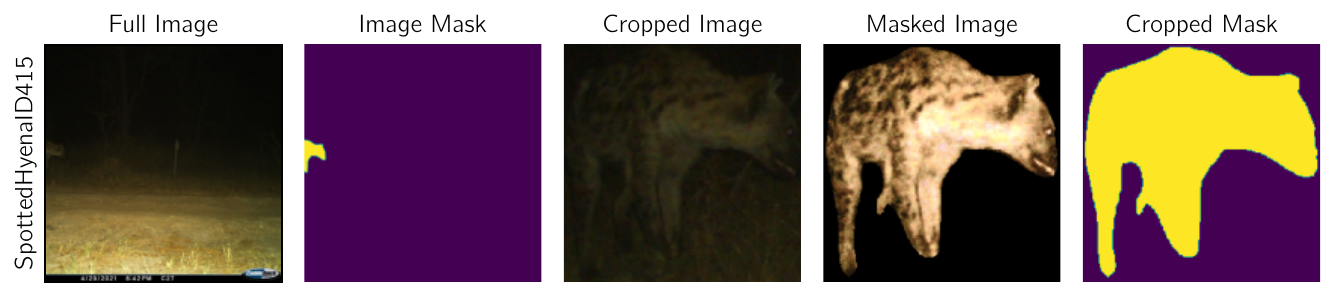} \\

        \caption{Sample images from each dataset. The impact of the preprocessing steps is visualized along row. The final preprocessed image is provided in the right-most column.}
    \end{subfigure}
    \caption{Quantitative and qualitative overview of our three animal ReID datasets. (a–c) Sighting frequency distributions for LeopardID102, SpottedHyenaID109, and SpottedHyenaID415 datasets, highlighting the long-tailed imbalance in observations per individual. (d) Sample images illustrating our preprocessing pipeline for each dataset.}
    \label{fig:dataset}
\end{figure*}

As part of this work, we release three challenging animal ReID datasets from camera trap surveys conducted in Zimbabwe: two contain spotted hyenas (SpottedHyenaID415 and SpottedHyenaID109), and one includes Leopards (LeopardID102). The images were obtained from a set of paired static camera traps distributed throughout survey sites. The images vary considerably in quality, time of day, viewpoint of the animal, and are highly unbalanced in terms of the number of images per individual. The timestamps and relative locations between the cameras are known. The SpottedHyenaID415 and LeopardID102 datasets originated from the same survey and contain all images of each species captured during the survey period. The SpottedHyenaID109 dataset is from a separate survey and comprises a subset of identity folders, as the full dataset contained more than 800 IDs with considerable uncertainty in label completeness (eg. potential missed matches). We therefore manually selected a subset of IDs for which we were confident in the individual classifications. The SpottedHyenaID109 dataset also contains hand-labelled binary orientation tags, indicating the visible side of the animal (left/right). 
All three datasets were collected directly by the authors of this work with permission from the landowners. Surveys were conducted by the property management organisations in partnership with the Wildlife Conservation Research Unit. No wildlife capture, handling, or invasive procedures were undertaken, and no permits or ethical approvals were required for the collection of these camera-trap images.
The released location data was anonymized by converting GPS coordinates into relative camera distances and removing image regions containing location-identifying features such as text. These privacy measures ensure the safety of local wildlife and nearby communities. We show dataset examples and statistics in \Cref{fig:dataset,tab:dataset_summary}. 

\begin{figure}[t]
	\centering
    \includegraphics[width=1\textwidth]{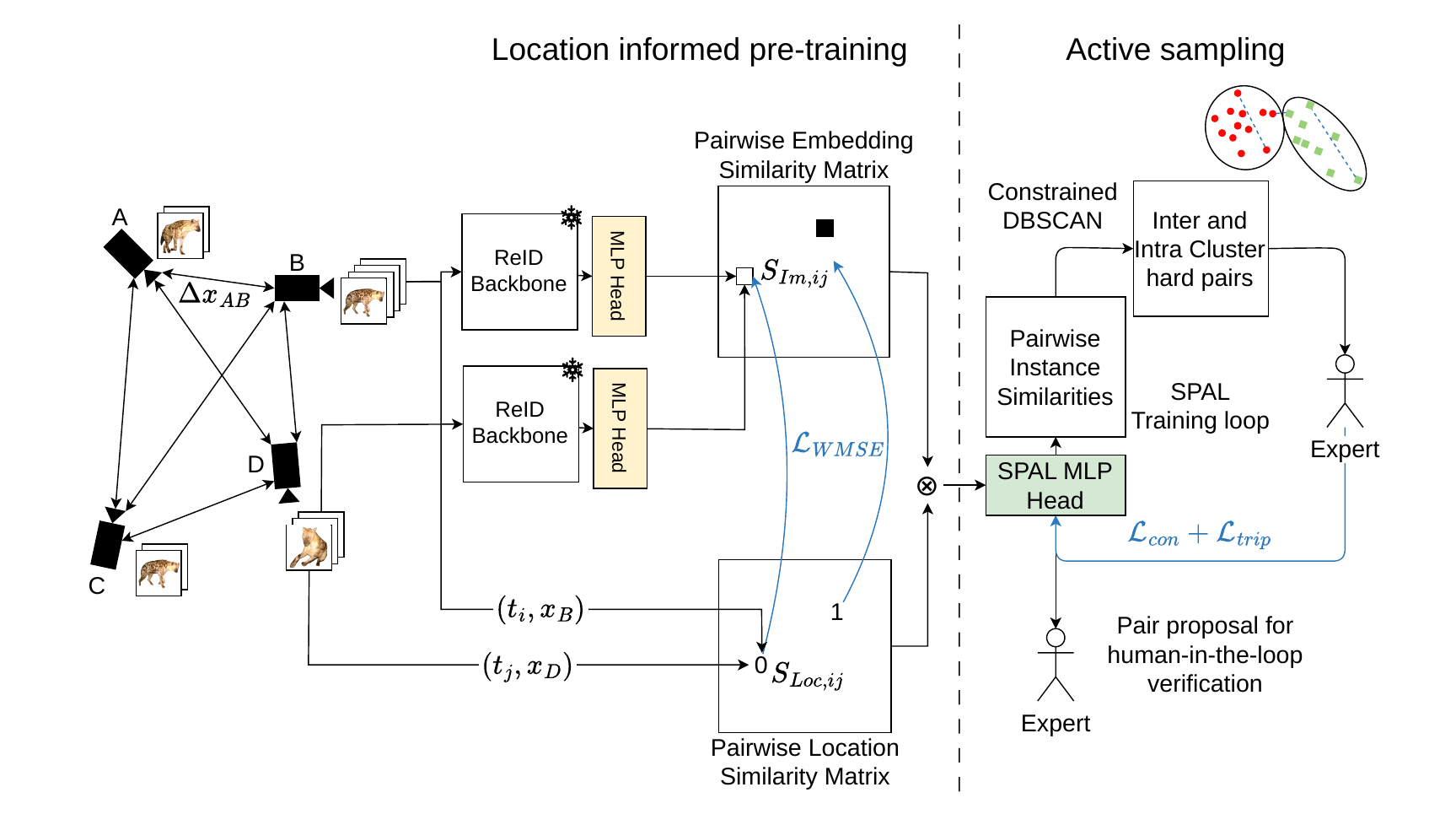}
    \caption{Overview of the proposed \net~framework. Camera-trap images are encoded using a frozen ReID backbone and refined via a network head trained on location-derived pseudo-labels. The adapted visual similarities are fused with spatio-temporal feasibility scores and used within an active sampling loop for efficient human-in-the-loop labeling.}
	\label{fig:system-diagram}
\end{figure}

\begin{table}[!htbp]
\centering
\caption{Statistics of our animal ReID datasets. Labelled images refer to images with expert-identified animal id.}
\label{tab:dataset_summary}
\begin{tabular}{|l|c|c|c|c|}
\hline
\textbf{Dataset} &
\textbf{Species} &
\textbf{\# Individuals} &
\textbf{\# Labeled Images} &
\textbf{\# of Cameras} \\
\hline
LeopardID102      & Leopard & 102 & 717 & 105\\
SpottedHyenaID109     & Spotted Hyena & 109  & 704 & 125\\
SpottedHyenaID415 & Spotted Hyena & 415 & 1871 & 105 \\
\hline
\end{tabular}
\end{table}

\section*{Methods}
\subsection*{Preprocessing}
We apply histogram equalization and histogram matching as preprocessing steps to reduce the major variations in image color and intensity. These image transforms minimize the impact of low-level lighting variations on the downstream ReID task. Thus, we visually select a reference image per dataset for histogram matching. Next, we use grounded SAM2~\cite{ren2024groundedsamassemblingopenworld} to mask and crop the individual animals as shown in \Cref{fig:dataset}.

\subsection*{Spatio-Temporal Model}

The spatio-temporal feasibility score is based on the straight-line speed between two timestamped detections taken at known camera locations. This straight-line speed represents the minimum travel speed required for two detections to correspond to the same individual. As such, an excessively high minimum speed indicates the spatio-temporal plausibility of the association. For each image pair, the euclidean distance between their respective camera trap location $\mathbf{x_i}$ and $\mathbf{x_j}$ and the time difference between their timestamps $t_i$ and $t_j$, are computed and normalised to a range between 0 and 1 (\(\Delta \tilde x\) and \(\Delta \tilde t\)) as detailed in \cref{eq:delta_x,eq:delta_t}. This normalization ensures scale invariance across survey sites and reduces the scoring function to a single tunable hyperparameter $s$ with exponential decay applied to the normalized pairwise speeds (Eq. \ref{eq:s_loc}).


\begin{align}    
    &\Delta x_{ij} = \left\lVert \mathbf{x_i}-\mathbf{x_j} \right\rVert_2, \ \ \Delta \tilde{x}_{ij}=\frac{\Delta x_{ij} - \Delta x_{min}}{\Delta x_{max} - \Delta x_{min}}\label{eq:delta_x}\\
    &\Delta t_{ij} = |t_i-t_j|, \ \ \Delta \tilde{t}_{ij}=\frac{\Delta t_{ij} - \Delta t_{min}}{\Delta t_{max} - \Delta t_{min}}\label{eq:delta_t}\\
    &S_{loc}(i, j)=\exp\biggl(-s\frac{\Delta\tilde{x}_{ij}}{\Delta\tilde{t}_{ij}}\biggl)\label{eq:s_loc}
\end{align} 

\begin{figure}[t]
	\centering
	\includegraphics[width=0.75\textwidth]{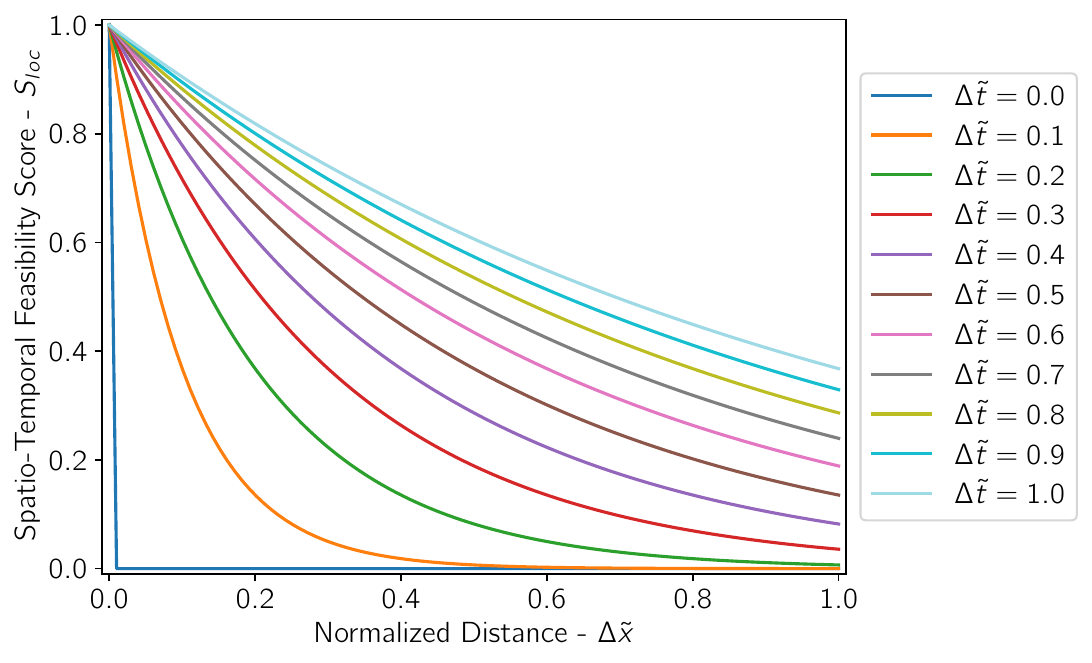}
    
    \caption{Spatio-temporal feasibility score as a function of normalised distance $\Delta \tilde{x}$. shown for multiple values of $\Delta \tilde{t}$. Larger $\Delta \tilde{t}$ results in slower decay with distance, reflecting increased tolerance to spatial displacement over longer time intervals.} 
	\label{fig:loc-score}
\end{figure}

This distribution biases towards higher match feasibility towards spatially close detections, even when the temporal separation \(\Delta \tilde t\) is large, as seen in \Cref{fig:loc-score}. This bias is desirable because it implicitly encodes the notion of home range which describes the phenomena that individual animals tend to remain within a limited spatial region even over extended periods \cite{10.1644/11-MAMM-S-177.1}.
The strength of this spatial bias is controlled by the parameter \(s\), which governs the rate at which the score decays with increasing speed. The parameter \(s\) was set to 1 for the survey sites following a grid search and held fixed across experiments. A spatio-temporal score of zero indicates a strong negative constraint, corresponding to image pairs that are highly unlikely to depict the same individual, e.g., images are captured at similar times but at distant locations. These trends can be seen in \Cref{fig:loc-score} when $\Delta{\tilde{t}}$ is near zero. In contrast, matching scores close to one indicate proximity and act as a weak positive signal. We define these pairs as \textit{weak} because intra-identity correspondence is not guaranteed. Such pairs may correspond to either the same individual or to different individuals within the same vicinity, including pairs within the same social pack.

\subsection*{Image Model}

We make use of a frozen pre-trained animal Re-ID model as our visual backbone. To emphasise the image-model agnostic nature of our approach, we test with both the MiewID~\cite{miewid} and MegaDescriptor~\cite{Cermak_2024_WACV} multi-species ReID models in \net. On top of this backbone, we train a lightweight MLP \(f_{\theta}\) consisting of a down-projection to a lower-dimensional space followed by an up-projection back to the original embedding dimension. The head is implemented using two fully connected layers with a low-rank bottleneck and a residual connection. The residual connection adds the head's output to the original embedding, allowing the model to learn incremental adjustments to the pre-trained features. We select this lightweight design as it reduces the number of trainable parameters and helps mitigate overfitting in the presence of limited and highly imbalanced training data.

We derive the training supervision from the computed spatio-temporal feasibility score as explained in subsection \textit{Spatio-Temporal Model}. The set of all image pairs with high feasibility scores (\(S_{loc}(i, j)=1\)) are treated as weak positive samples ($\mathcal{P}^+$), while the set of all image pairs with low feasibility scores (\(S_{loc}(i,j)=0\)) are treated as confident negative samples ($\mathcal{P}^-$). During training, we construct a pairwise similarity matrix \(S_{im}(i,j)\) from the MLP head embeddings using cosine similarity and clip negative values. We apply a weighted L2 loss \cref{eq:image_loss} to align the image embedding similarities with the spatio-temporal feasibility scores for the selected pairs (all image pairs with \(S_{loc}=1\) and \(S_{loc}=0\)). We propose that low feasibility pairs provide a stronger constraint than high feasibility pairs which do not guarantee identity correspondence.

\begin{align}
&\mathcal{L}_{WMSE}
=\frac{\lambda_p}{|\mathcal{P}^+|}\sum_{(i,j)\in\mathcal{P}^+}\left(S_{im}(i,j)-1\right)^2
+
\frac{\lambda_n}{|\mathcal{P}^-|}\sum_{(i,j)\in\mathcal{P}^-}\left(S_{im}(i,j)\right)^2
\label{eq:image_loss}\\
&\mathcal{P}^+=\{(i, j) \mid S_{loc}(i, j)=1\}\label{eq:pos_pair_set}\\
&\mathcal{P}^-=\{(i, j) \mid S_{loc}(i, j)=0\}\label{eq:neg_pair_set}
\end{align}

This objective encourages similar embeddings for spatio-temporally proximate samples while penalizing similarities for impossible pairs. Although high feasibility scores do not guarantee identity correspondence, they frequently capture co-occurrence within the same social group or pack, which gives a strong prior for the downstream task. Further, we include weakly positive pairs during training to prevent representational collapse, as training exclusively on negative pairs would encourage the model to trivially minimize similarity across all samples. As the goal of the system is to recommend candidate matches for human annotation rather than perform fully automatic identification, exploiting these weak positive cues improves the relevance of the proposed image pairs. Formulating the learning objective in a pairwise manner converts the problem into a binary classification task (intra-individual vs. inter-individual pairs), which increases the number of training signals per label, removes the need for prior knowledge of the number of identities, and partially alleviates class imbalance. Final pairwise similarities are computed through element-wise multiplication of spatio-temporal and image embedding similarities, yielding scores that leverage both modalities and improve robustness over image-only baselines.

\subsection*{Active Pair Sampling}
The objective of the annotation stage is to minimize the total number of required pairwise comparisons pursued by a human annotator. Expert annotated pairwise labels are encoded as constraints: confirmed matches are treated as \textit{must-link} constraints, while confirmed non-matches between distinct individuals are treated as \textit{cannot-link} constraints. At each annotation step, these pairwise constraints are propagated to infer all implicit relationships that can be derived from the new pairwise label obtained by the annotator. A simple greedy strategy, which always proposes the highest scoring image pair for labeling, is effective at identifying easy matches early but diminishes in performance with an increasing number of queries. As annotation progresses, only the ambiguous low-confidence set of matching image pairs remains for which the pairwise score quality is uninformative regarding their labels. Since we obtain ground-truth pairwise labels when the human annotator labels queried pairs and their labels are propagated, these annotations can be directly reused as a supervisory signal. To address this, we adapt the uncertainty-based pair sampling and training strategy introduced in the SPAL method\cite{jin2022fewerlabelssupportpair}. We use our pairwise similarity scores to train an additional two-layer MLP, inspired by the training methodology presented in SPAL \cite{jin2022fewerlabelssupportpair}. Uncertain pairs are identified by clustering the data with constrained DBSCAN \cite{jin2022fewerlabelssupportpair,cons-dbscan}, selecting the most dissimilar pair of images within each cluster and the most similar pair of images between each cluster. These pairs are passed to the expert for labeling and the annotations serve as a training signal. In accordance with SPAL, we apply a combination of triplet and contrastive losses to push samples towards cluster centroids. SPAL aims to select informative pairs for annotation by sampling pairs from cluster boundaries. However, as clusters begin to stabilize, fewer new samples are selected as the cluster boundaries become static, at which point we propose to extend SPAL with a confidence-based sampling scheme to select the highest-scoring pair for labeling, which we refer to as ConfSPAL. We demonstrate that dynamic sampling strategies can effectively combine informativeness with coverage, outperforming prior approaches that prioritize informative samples exclusively in terms of queried human comparisons.

\section*{Experimental Results}
\subsection*{Experimental Settings}
During the location-informed pre-training stage, we use the Adam optimizer with a learning rate of $1e-3$ and train for 300 epochs with full-batch updates. The L2 loss weights, $\lambda_n$ and $\lambda_p$, are empirically set to 2 and 1, respectively. Additionally, we train the active sampling head during the annotation stage separately using an Adam optimizer at a learning rate of $3e-4$ with full batch updates. For constrained DBSCAN clustering, the minimum number of samples was set to $2$. The DBSCAN neighbourhood radius $\varepsilon$ was selected through a grid search by qualitative inspection of the resulting distribution of cluster sizes, with the objective of limiting excessively large clusters, typically capped to the range of $10$-$50$ samples, while reducing the number of single-element clusters. The selected values of $\varepsilon$ for the LeopardID102, SpottedHyenaID109, and SpottedHyenaID415 were 0.44, 0.44 and 0.49 for \net~(MD), 0.62, 0.60, and 0.57 for \net~(MI), 0.21, 0.22, and 0.19 for MegaDescriptor, and 0.47, 0.45, and 0.41 for MiewID, respectively.
\begin{figure}[htp]

\centering
\begin{subfigure}{.33\textwidth}
    \centering
    \includegraphics[width=1\textwidth]{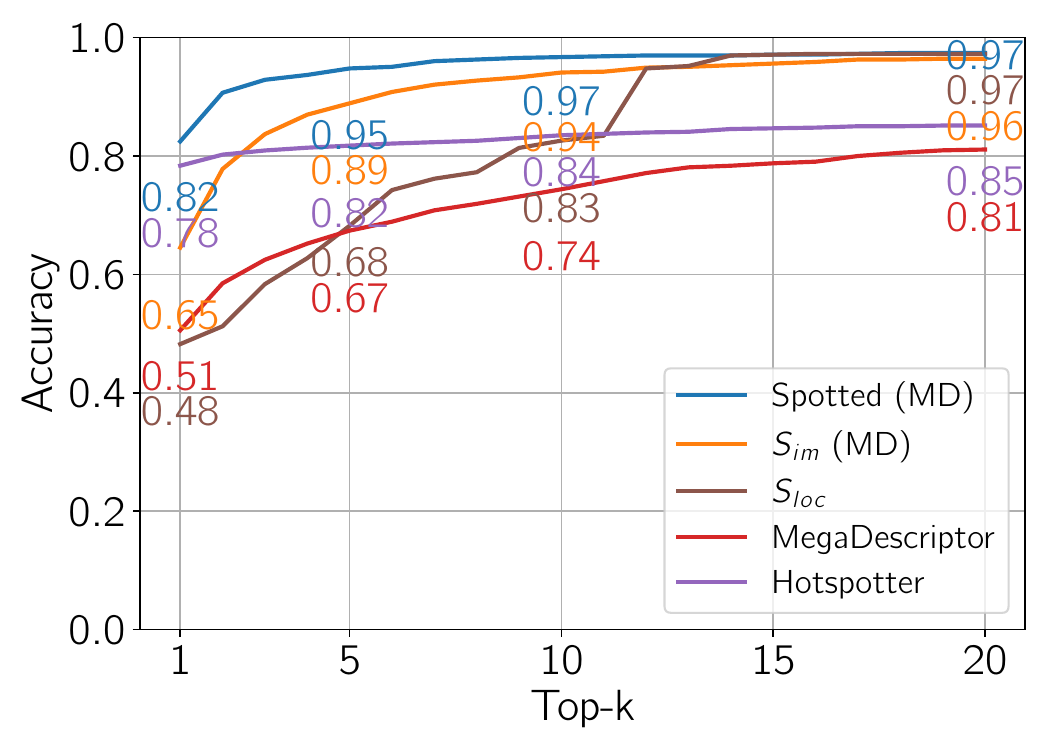}\hfill
    \caption{LeopardID102}
\end{subfigure}
\begin{subfigure}{.33\textwidth}
    \centering
    \includegraphics[width=1\textwidth]{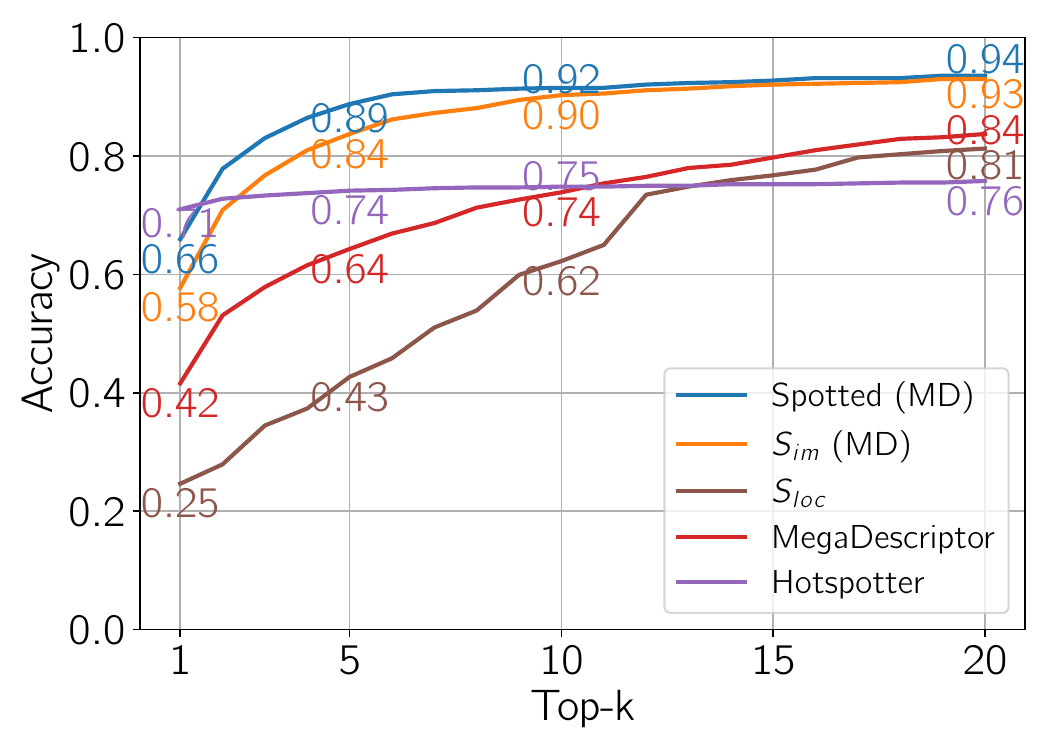}\hfill
    \caption{SpottedHyenaID109}
\end{subfigure}
\begin{subfigure}{.33\textwidth}
    \centering
    \includegraphics[width=1\textwidth]{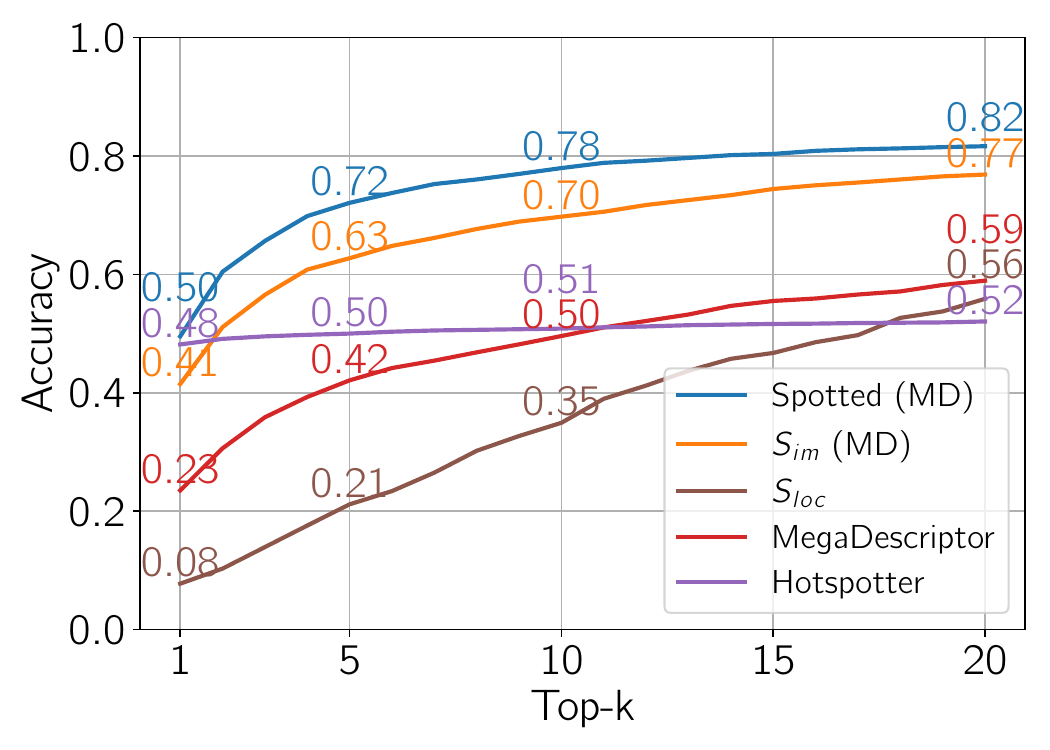}
    \caption{SpottedHyenaID415}
\end{subfigure}
\centering
\begin{subfigure}{.33\textwidth}
    \centering
    \includegraphics[width=1\textwidth]{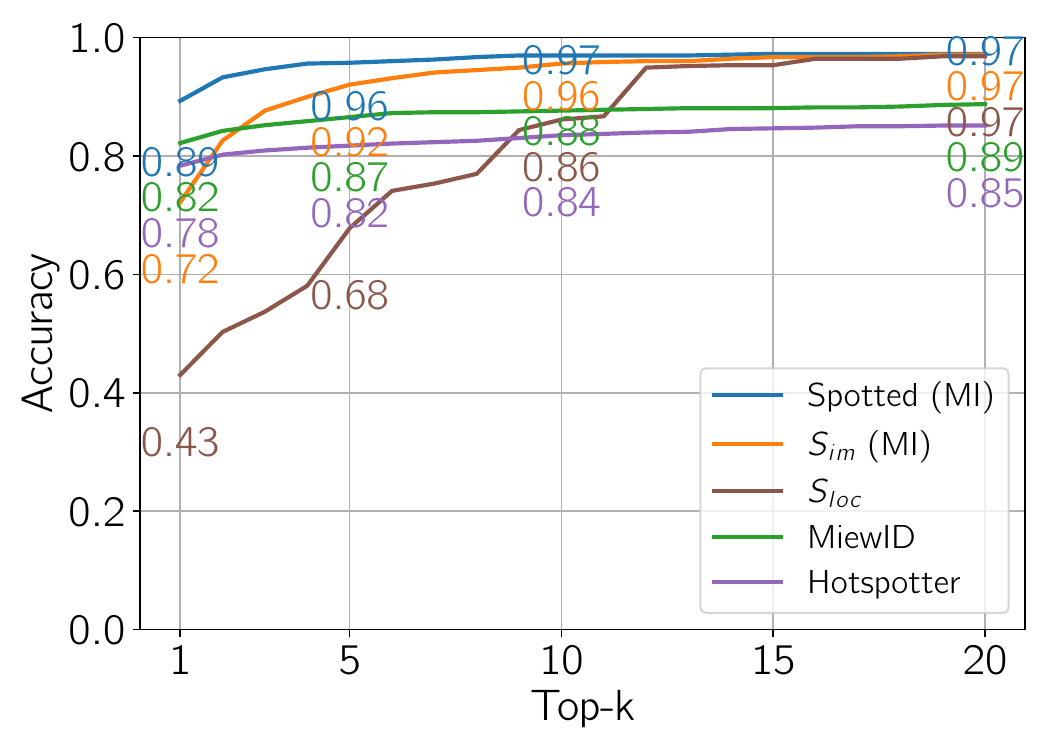}\hfill
    \caption{LeopardID102}
\end{subfigure}
\begin{subfigure}{.33\textwidth}
    \centering
    \includegraphics[width=1\textwidth]{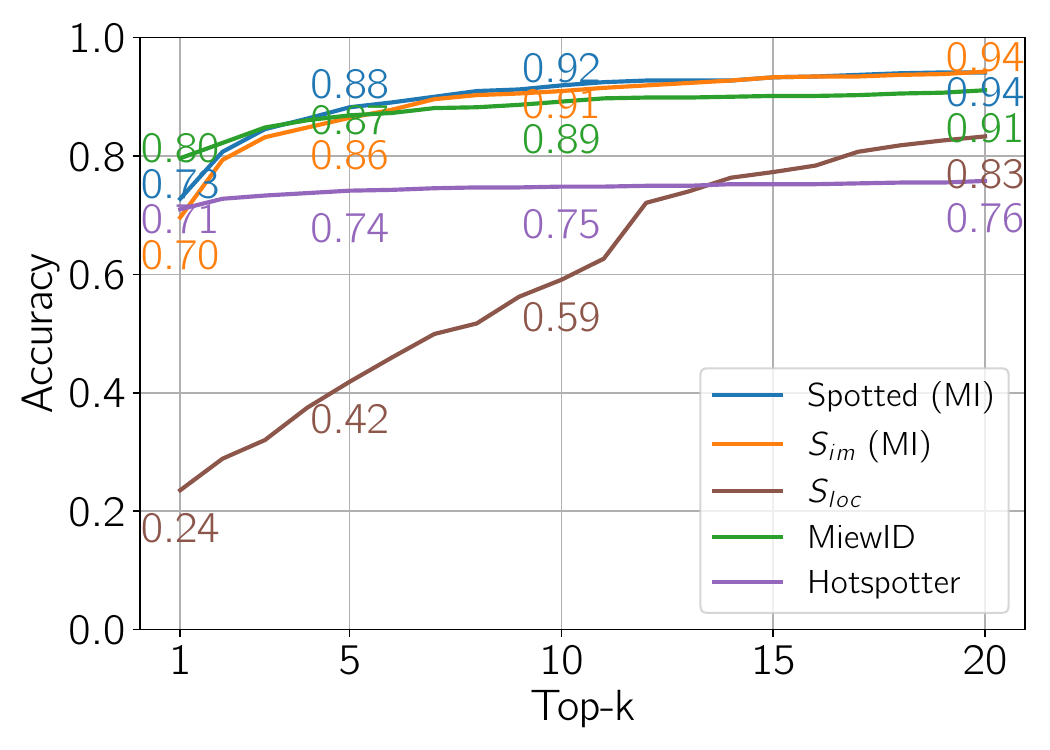}\hfill
    \caption{SpottedHyenaID109}
\end{subfigure}
\begin{subfigure}{.33\textwidth}
    \centering
    \includegraphics[width=1\textwidth]{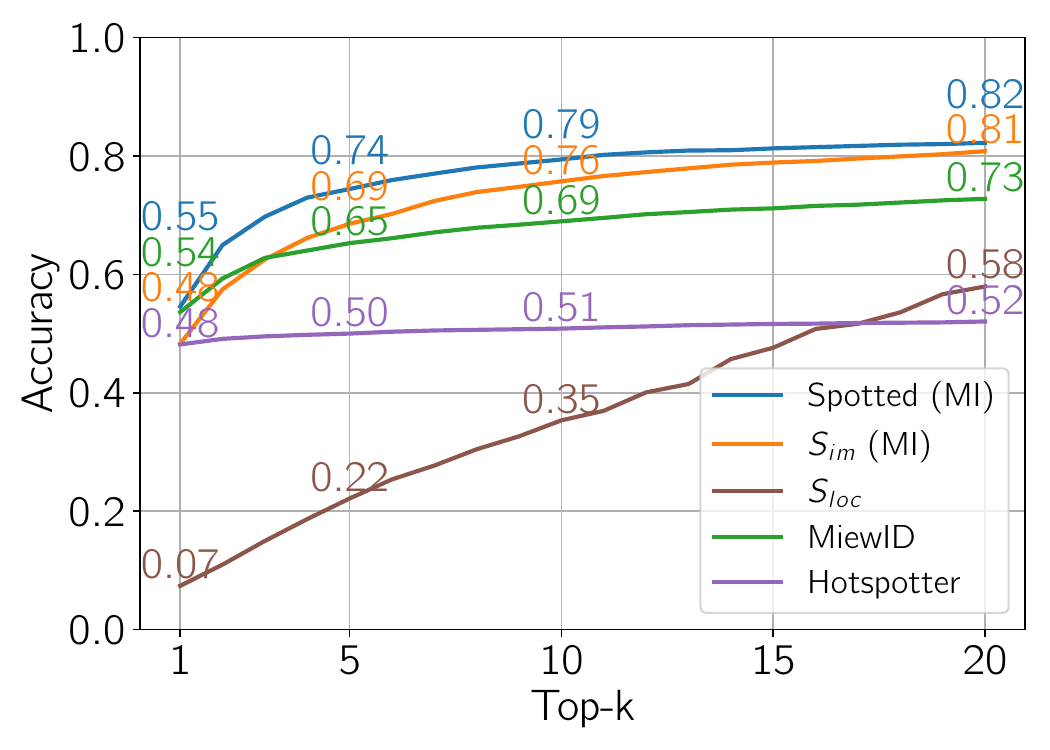}
    \caption{SpottedHyenaID415}
\end{subfigure}

\caption{Top-k accuracy curves for our three datasets. The top-k accuracy is computed as the proportion of query images for which at-least one true match is retrieved within the top-k highest scoring image pairs. (a-c) MegaDescriptor is used as the frozen image backbone. (d-f) MiewID is used as the frozen image backbone.}
\label{fig:topk-acc}

\end{figure}

We evaluate our approach against three baselines, HotSpotter \cite{crall2013hotspotter}, MegaDescriptor \cite{Cermak_2024_WACV}, and MiewID \cite{miewid}. MegaDescriptor and MiewID represent the current state-of-the-art in multi-species animal ReID and thus serve as strong baselines, while HotSpotter is still widely used in practical wildlife monitoring pipelines. We evaluate our model in two stages. First, we assess the quality of our pairwise similarity scores using top-$k$ identification accuracy. Second, we simulate a human-in-the-loop dataset labeling scenario and measure annotation efficiency in terms of the number of queried image pairs until all positive pairwise matches are discovered. 

\subsection*{Scoring Performance}
As shown in \Cref{fig:topk-acc}, \net~outperforms visual-only baselines across all datasets. In comparison to MegaDescriptor embeddings, our method achieves improvements of 28pp, 24pp, and 31pp over this backbone and 9pp, 2pp and 9pp over MiewID in top-5 accuracy on the LeopardID102, SpottedHyenaID109, and SpottedHyenaID415 datasets. While Hotspotter remains competitive in terms of top-1 accuracy, its performance stagnates for larger k, reflecting  its tendency to identify a small number of strong matches and missing further plausible candidate matches. 

To explore the relative impact of the spatio-temporal prior compared to the visual features, we additionally report the top-k accuracy curves for the image only (\(S_{im}\)) and location only (\(S_{loc}\)) scores. Although the spatio-temporal score alone shows limited accuracy at low k values, it achieves competitive top-k accuracy for high k values. In particular, we observe top-20 accuracies of 97\%, 81\%, and 56\% for the LeopardID102, SpottedHyenaID109, and SpottedHyenaID415 datasets, respectively. This shows that spatial and temporal data capture highly informative priors, helping to constrain the search space for plausible matches.
It is also important to note the impact of species and population size on the effectiveness of the spatio-temporal priors. For example, the SpottedHyenaID415 dataset, which contains $415$ individuals, exhibits substantially lower top-1 accuracy on location-only scores (8\%) compared to the SpottedHyenaID109 dataset (25\%) which contains 109 individuals (Table \ref{tab:dataset_summary}). This is expected, as a larger population leads to more individuals per location, reducing the discriminative power of location-only priors.

These results also reflect some differences in species movement behaviour. Although the LeopardID102 and SpottedHyenaID109 datasets contain a similar number of individuals ($102$ - $109$), the location-only model achieves substantially higher top-1 accuracy on LeopardID102 (48\%) than on SpottedHyenaID109 (25\%). This is in line with the movement patterns observed in the datasets, where leopards tend to remain in spatially consistent regions, whereas hyenas were more likely to travel longer distances between sightings, introducing greater uncertainty in location-based cues for hyenas. Consequently, we highlight the need to combine image and location-prior models to achieve the highest scores across all datasets and backbones. 

\begin{figure}[!htbp]
\centering

    \begin{subfigure}{.33\textwidth}
        \centering
        \clipbox{0pt 0pt 1.3cm 0pt}{%
            \includegraphics[width=\textwidth]{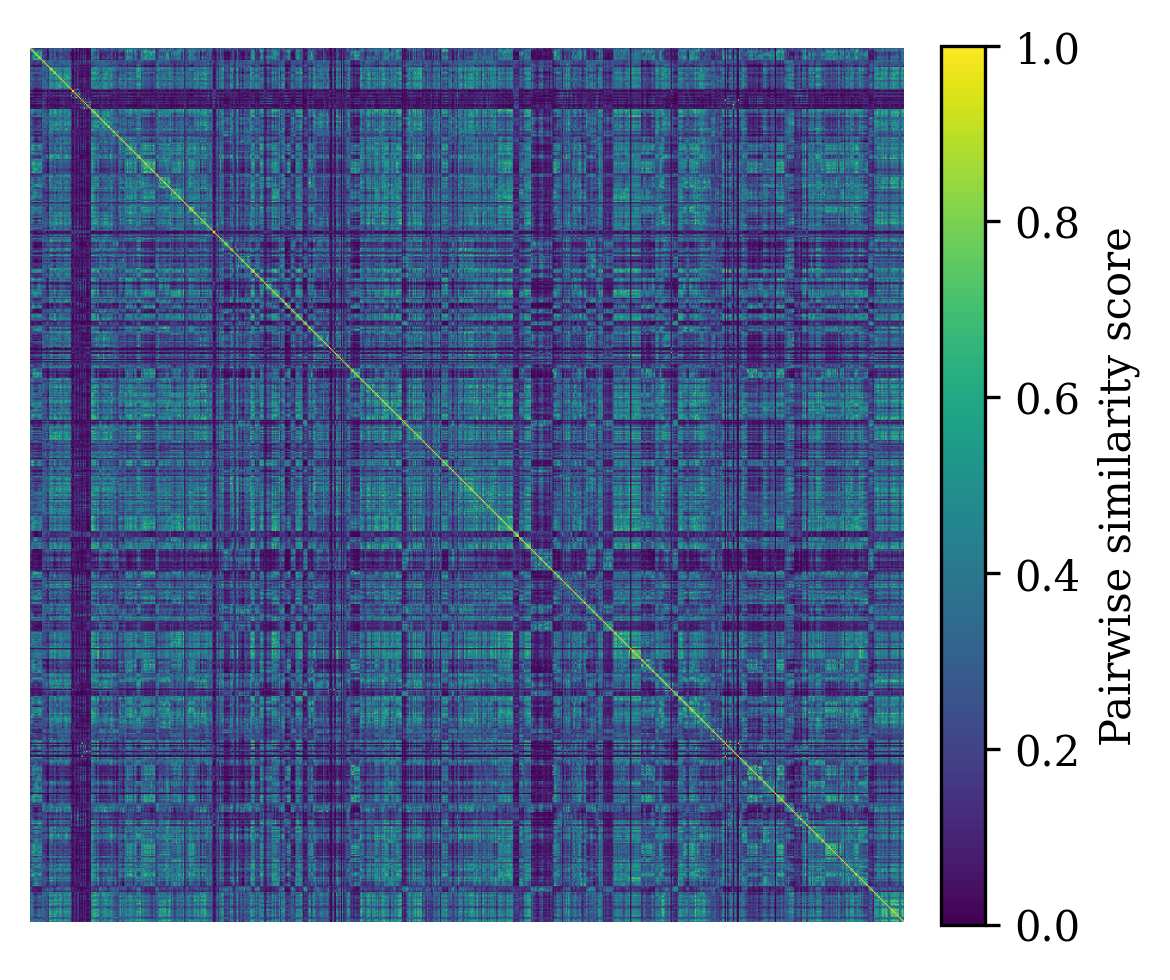}%
        }
        \caption{MegaDescriptor}
    \end{subfigure}\hspace{-2.5em}%
    \begin{subfigure}{.33\textwidth}
        \centering
        \clipbox{0pt 0pt 1.3cm 0pt}{%
            \includegraphics[width=\linewidth]{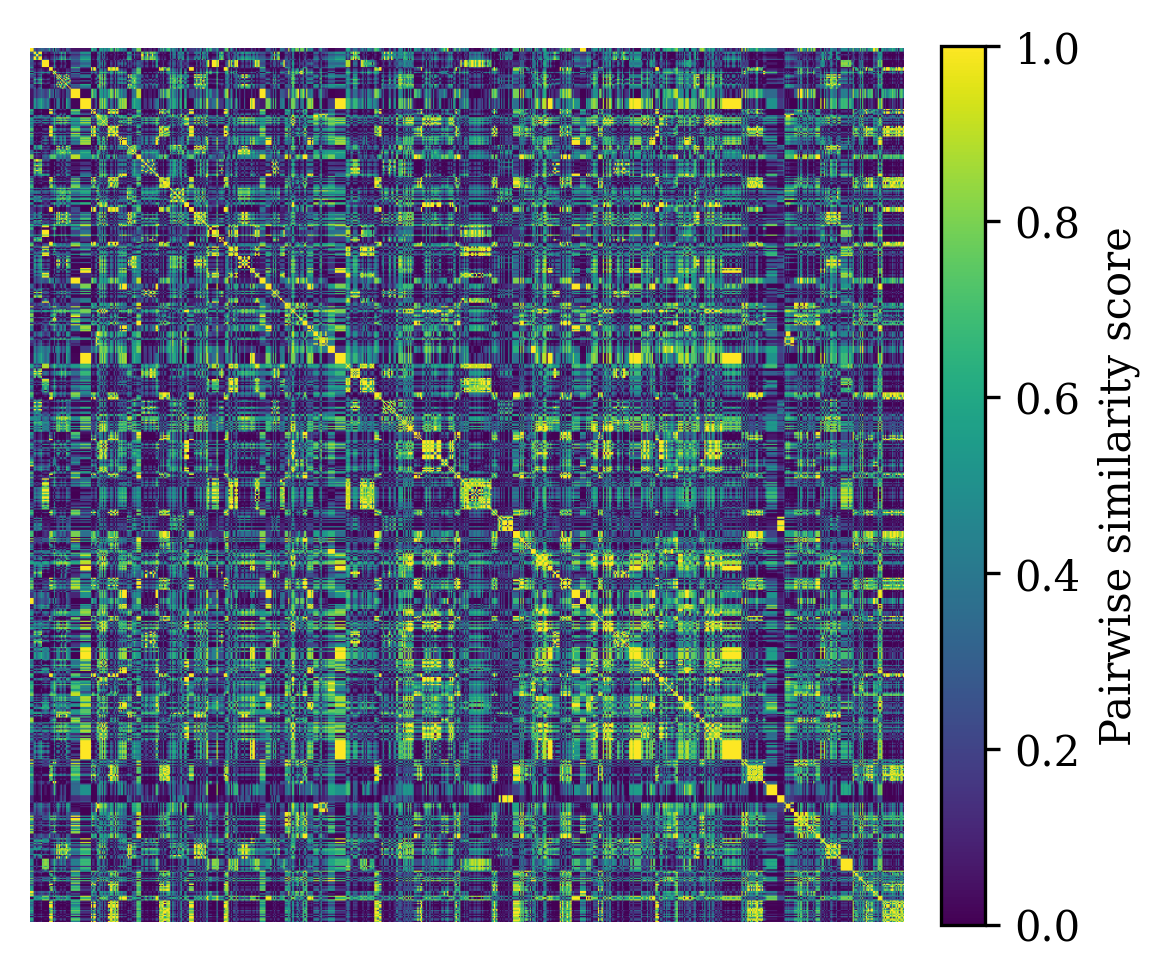}%
        }
        \caption{Spatio-temporal score - $S_{loc}$}
    \end{subfigure}\hspace{-2.5em}%
    \begin{subfigure}{.33\textwidth}
        \centering
        \clipbox{0pt 0pt 1.3cm 0pt}{%
            \includegraphics[width=\linewidth]{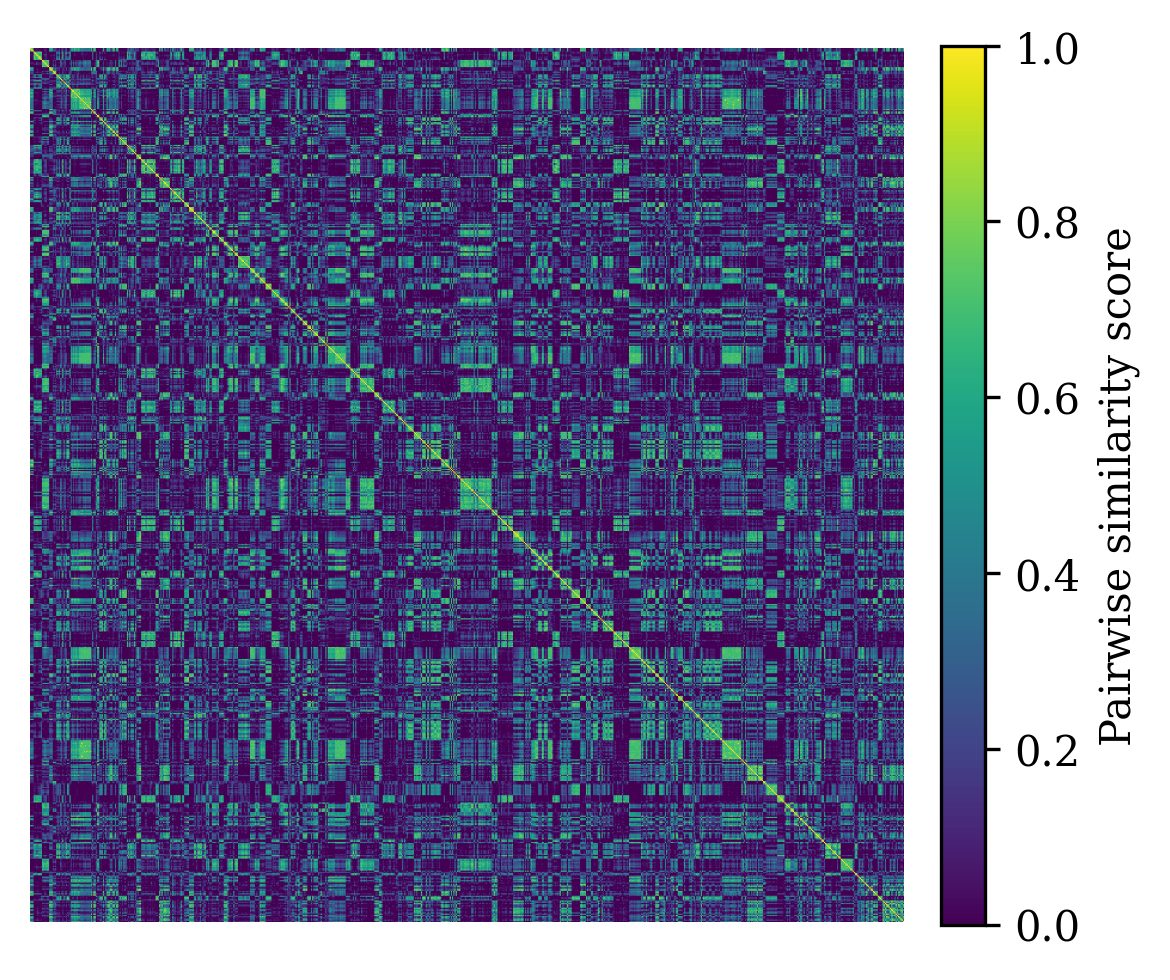}%
        }
        \caption{Adapted image embeddings - $S_{im}$}
    \end{subfigure}
    \begin{subfigure}{.32\textwidth}
        \centering
        \clipbox{0pt 0pt 1.3cm 0pt}{
        \includegraphics[width=\linewidth]{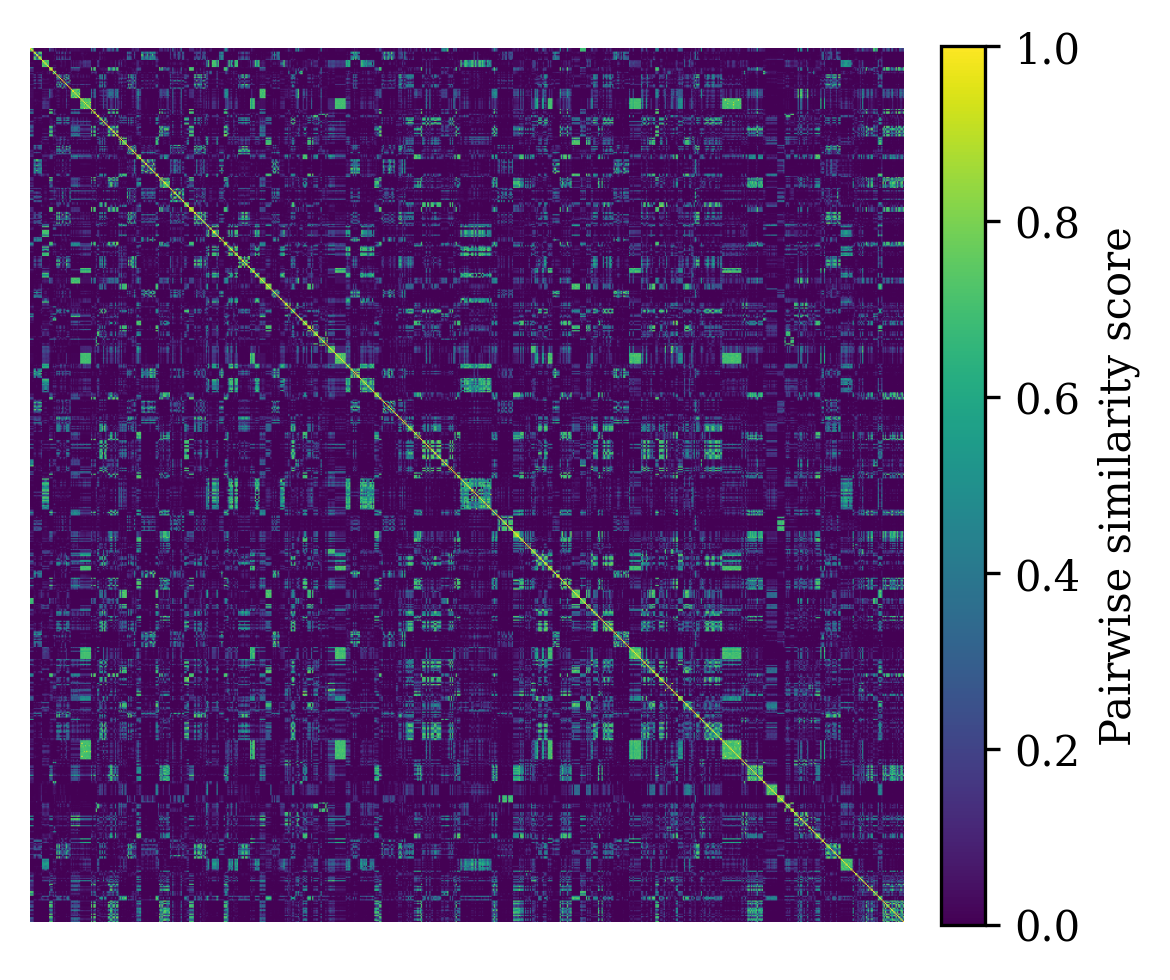}
        }
        \caption{\net{} - $S_{im} \odot S_{loc}$}
    \end{subfigure}\hspace{-2.5em}
        \begin{subfigure}{.32\textwidth}
        \centering
        \clipbox{0pt 0pt 1.3cm 0pt}{
        \includegraphics[width=\linewidth]{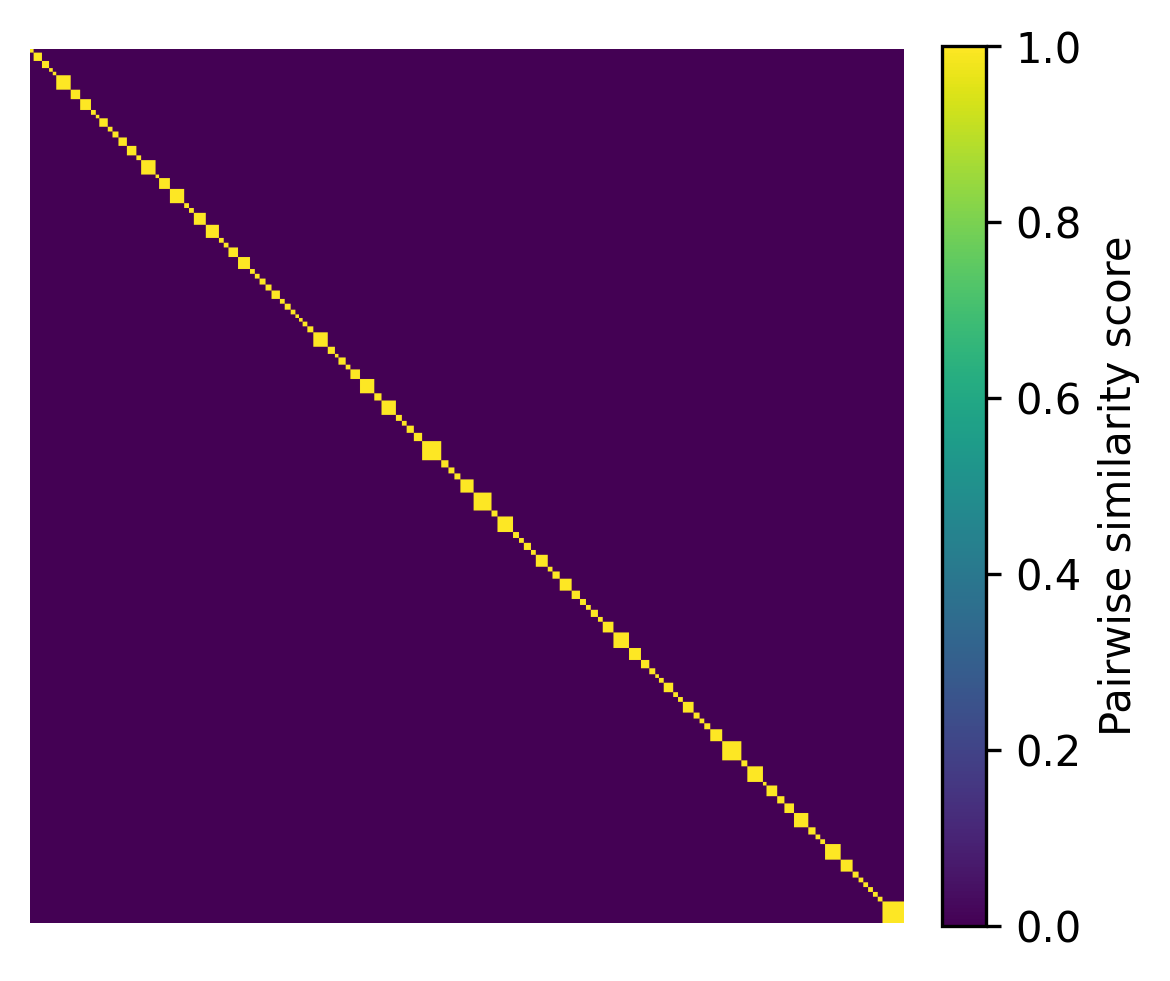}
        }
        \caption{Ground Truth}
    \end{subfigure}\hspace{-1.75em}
    \begin{subfigure}{.069\textwidth}
        \centering
        \includegraphics[width=\linewidth, trim=32.328cm 0 0 0, clip]{figs/GRZ_SH_CLEAN_ground_truth.png}
        \vspace{-0.1em}
    \end{subfigure}
\caption{Similarity matrices between all pairs of images within the SpottedHyenaID109 dataset. The data has been ordered by animal id.}
\label{fig:GRZ_SM}
\end{figure}

\Cref{fig:GRZ_SM} shows the pairwise similarity matrices for the SpottedHyenaID109 dataset at different stages of the proposed framework. The pairwise similarity matrix of the MegaDescriptor embeddings in \figref{fig:GRZ_SM}a) visually exhibits limited discriminative structure, with the similarity scores distributed too uniformly across both matching and non-matching pairs to reliably distinguish between them. Adapting the embeddings using spatio-temporal pseudo-labels (\figref{fig:GRZ_SM}b) suppresses a large fraction of these spurious similarities, as can be seen from the lower-scoring regions throughout the matrix. Nonetheless, a subset of negative pairs remains highly scored. The location-based similarity matrix  (\figref{fig:GRZ_SM}c) assigns high scores to many image pairs, reflecting the coarse but inclusive nature of the spatial prior, while still ensuring that true matching pairs consistently receive high scores. Since spatio-temporal scores of zero correspond to guaranteed non-matches, whereas high-scoring pairs provide only a weak positive signal based on proximity, the training process tends to avoid false-negative matches. Because of this bias in the spatio-temporal scoring function, both the adapted image embeddings and the spatio-temporal scores tend to produce more false positives than false negatives. This is particularly useful in a human-in-the-loop setting because missed matches can lead to an exhaustive manual search for a match. Consequently, true matches generally receive high scores in both modalities. Element-wise fusion of the image and spatio-temporal similarity signals combines these signals by preserving pairs that are consistently strong in both representations while suppressing spurious matches arising from visual ambiguity or spatial proximity alone. This joint bias toward retaining potential matches allows the fused score to reduce erroneous high similarities while maintaining strong responses for genuine matches, producing a clearer block-diagonal structure that closely aligns with the ground-truth pairwise labels (\figref{fig:GRZ_SM}e).

\subsection*{Human-in-the-Loop Performance}

\begin{figure}[!htbp]
    \centering
    \begin{subfigure}{.33\textwidth}
        \centering
        \includegraphics[width=1\textwidth]{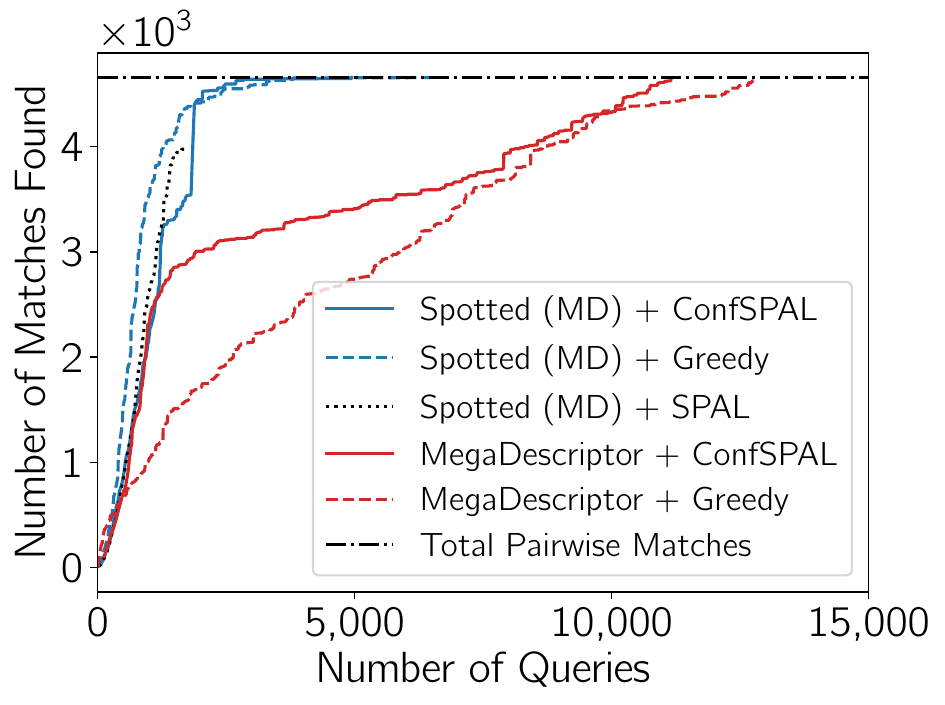}\hfill 
        \caption{LeopardID102}
    \end{subfigure}
    \begin{subfigure}{.33\textwidth}
        \centering
        \includegraphics[width=1\textwidth]{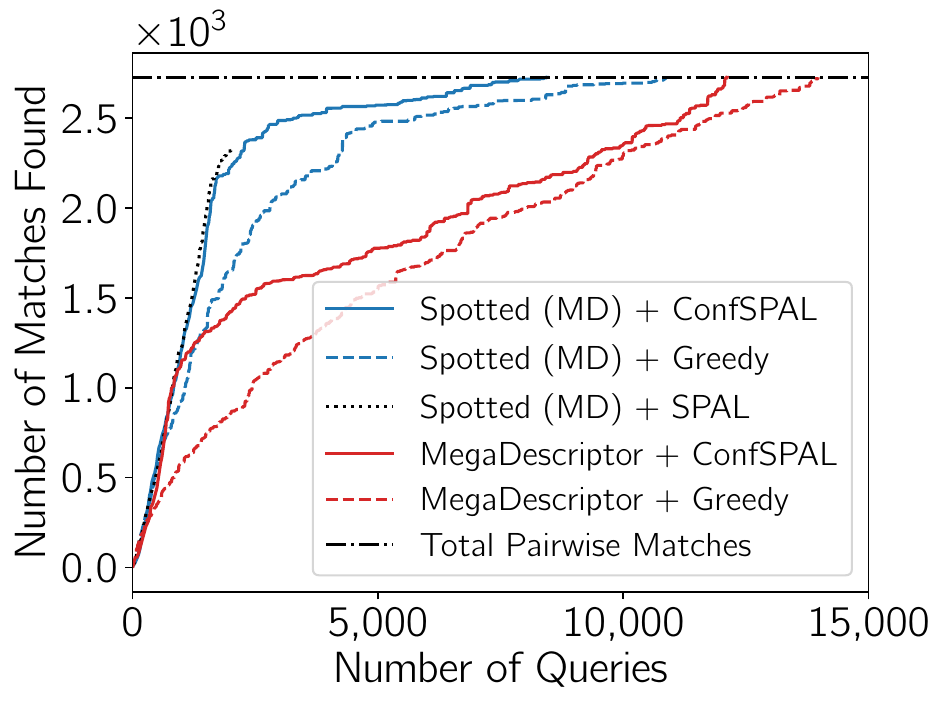}\hfill
        \caption{SpottedHyenaID109}
    \end{subfigure}
    \begin{subfigure}{.33\textwidth}
        \centering
        \includegraphics[width=1\textwidth]{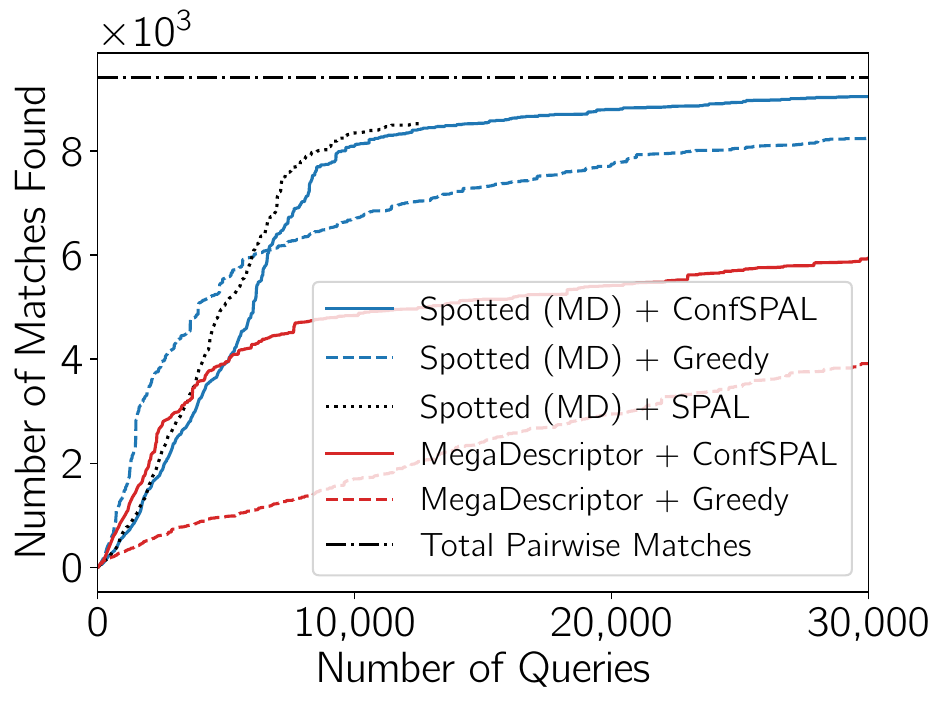}
        \caption{SpottedHyenaID415}
    \end{subfigure}
    \begin{subfigure}{.33\textwidth}
        \centering
        \includegraphics[width=1\textwidth]{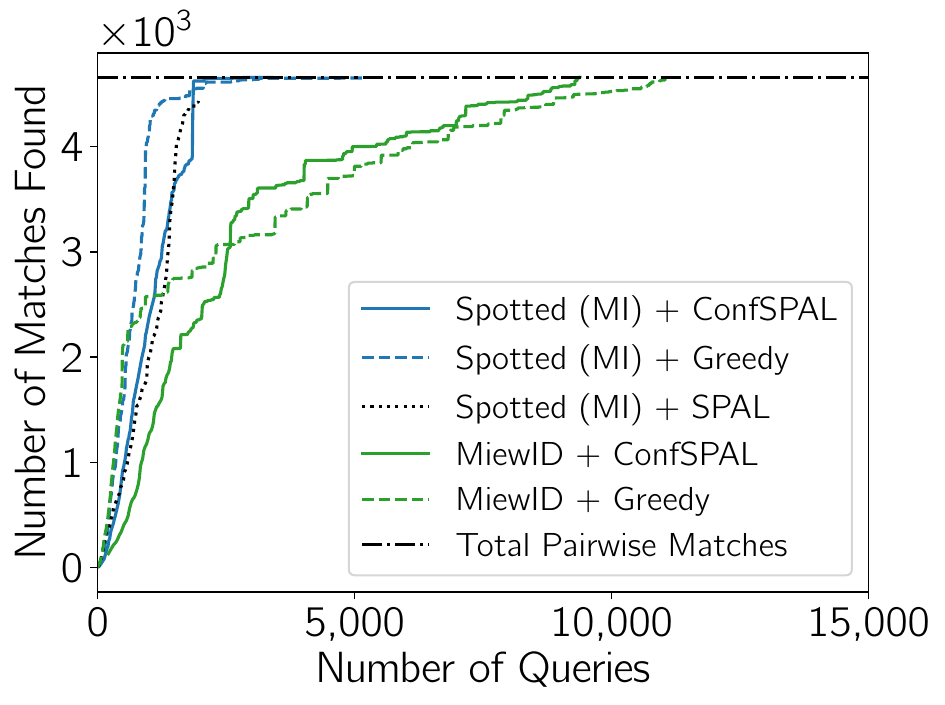}\hfill 
        \caption{LeopardID102}
    \end{subfigure}
    \begin{subfigure}{.33\textwidth}
        \centering
        \includegraphics[width=1\textwidth]{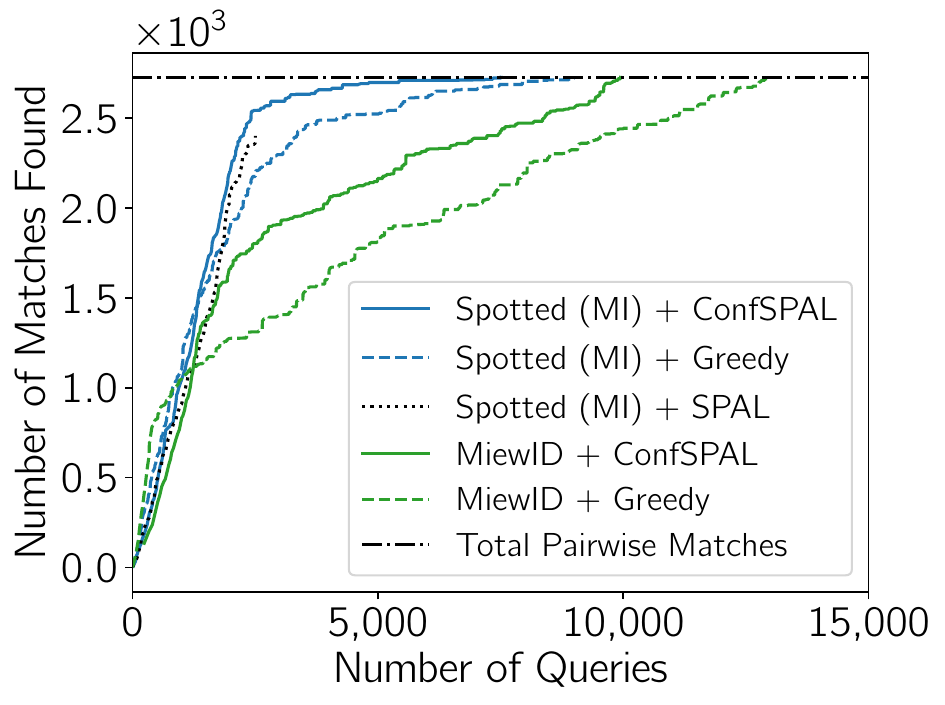}\hfill
        \caption{SpottedHyenaID109}
    \end{subfigure}
    \begin{subfigure}{.33\textwidth}
        \centering
        \includegraphics[width=1\textwidth]{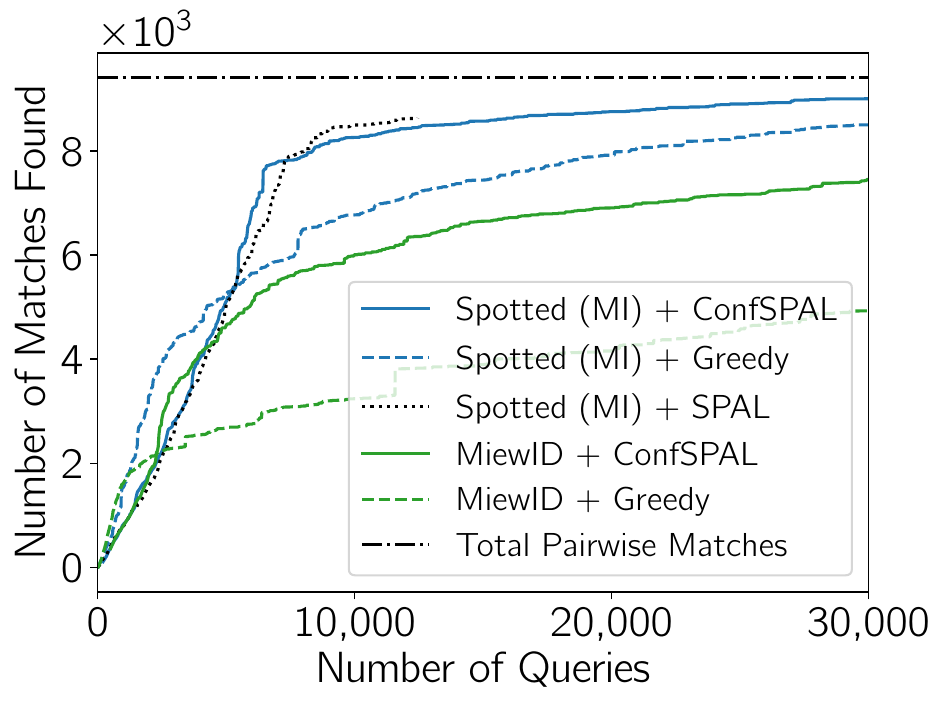}
        \caption{SpottedHyenaID415}
    \end{subfigure}
    
    \caption{Cumulative number of confirmed positive matches discovered as a function of annotation queries for our three datasets. We compare greedy sampling against our SPAL-inspired active sampling with and without the confidence sampling step. MegaDescriptor, MiewID and our \net~ similarities are compared. (a-c) MegaDescriptor (MD) is used as the frozen image backbone. (d-f) MiewID (MI) is used as the frozen image backbone. The dashed black line indicates the total number of positive pairwise matches.}
    \label{fig:queryvmatches}
\end{figure}

\begin{figure}[!htbp]{

    \centering
    \begin{subfigure}{.33\textwidth}
        \centering
        \includegraphics[width=1\textwidth]{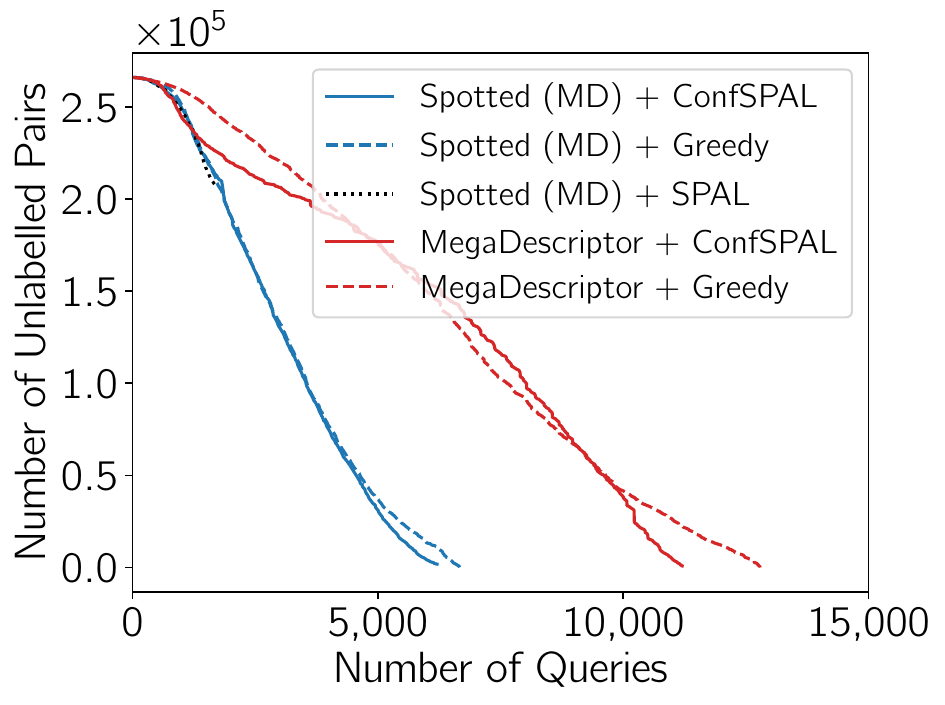}\hfill
        \caption{LeopardID102}
    \end{subfigure}
    \begin{subfigure}{.33\textwidth}
        \centering
        \includegraphics[width=1\textwidth]{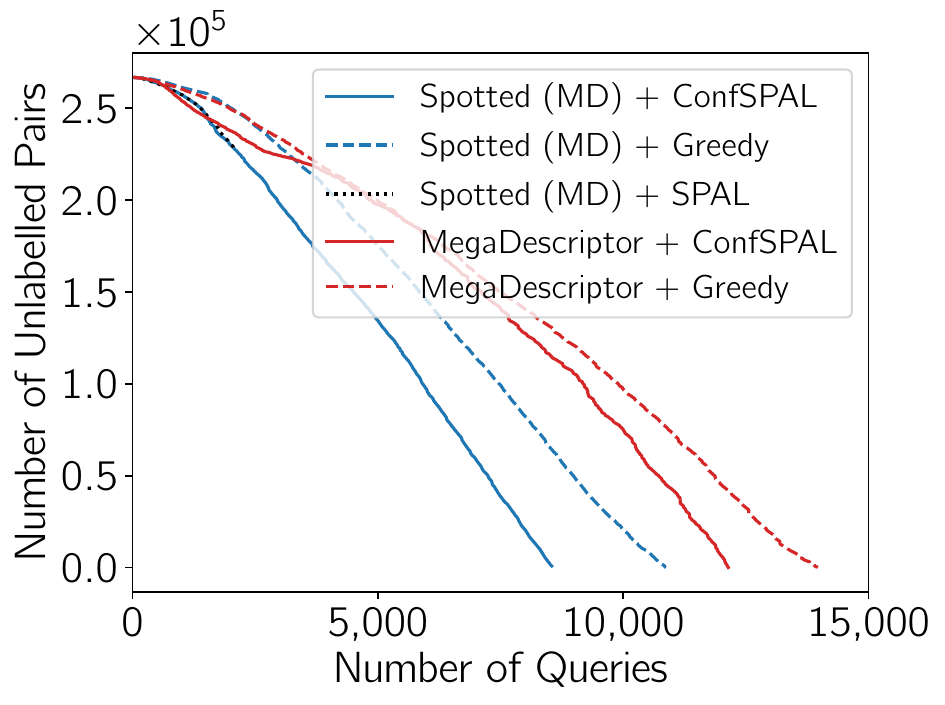}\hfill
        \caption{SpottedHyenaID109}
    \end{subfigure}
    \begin{subfigure}{.33\textwidth}
        \centering
        \includegraphics[width=1\textwidth]{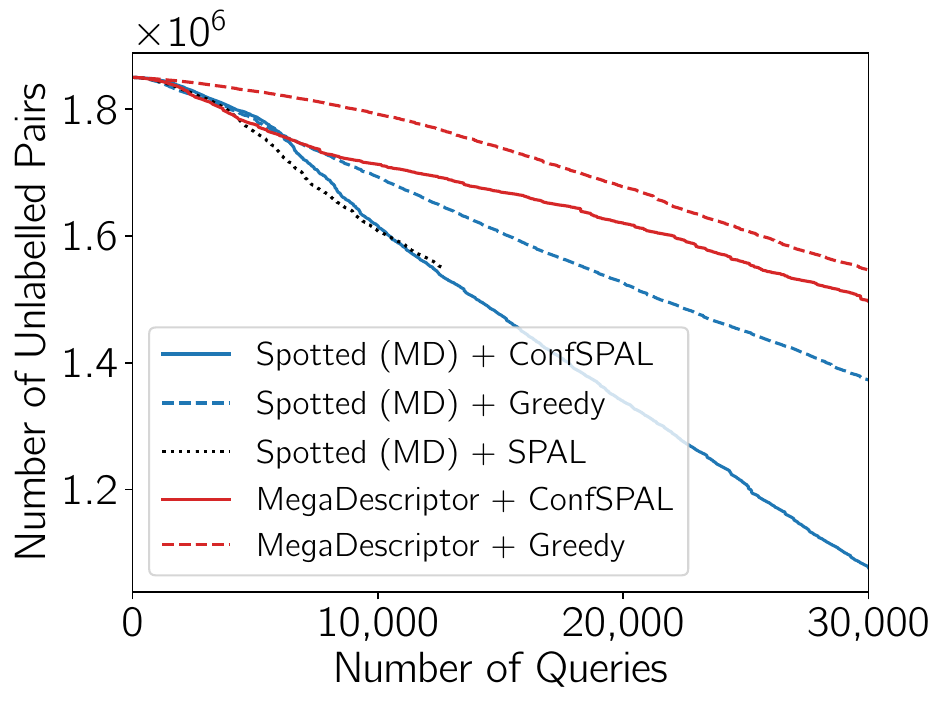}
        \caption{SpottedHyenaID415}
    \end{subfigure}
    \\
    \begin{subfigure}{.33\textwidth}
        \centering
        \includegraphics[width=1\textwidth]{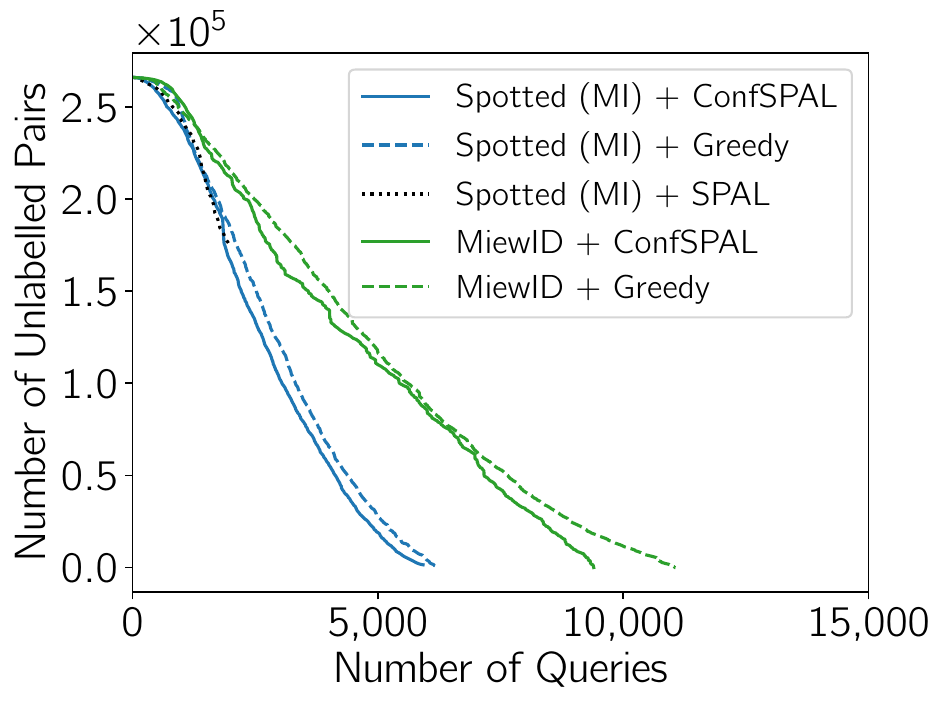}\hfill
        \caption{LeopardID102}
    \end{subfigure}
    \begin{subfigure}{.33\textwidth}
        \centering
        \includegraphics[width=1\textwidth]{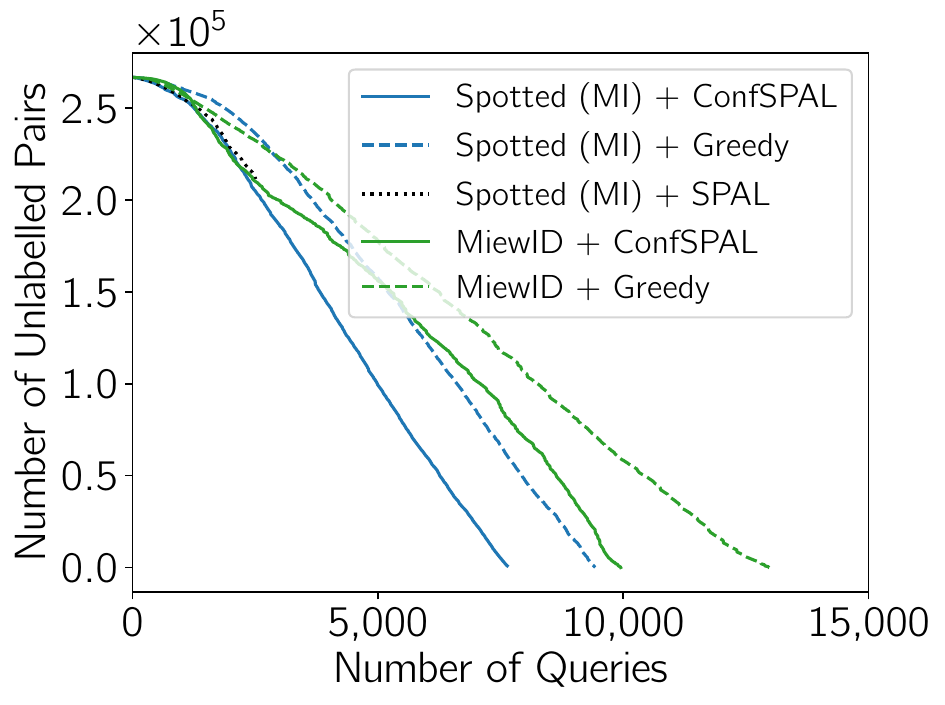}\hfill
        \caption{SpottedHyenaID109}
    \end{subfigure}
    \begin{subfigure}{.33\textwidth}
        \centering
        \includegraphics[width=1\textwidth]{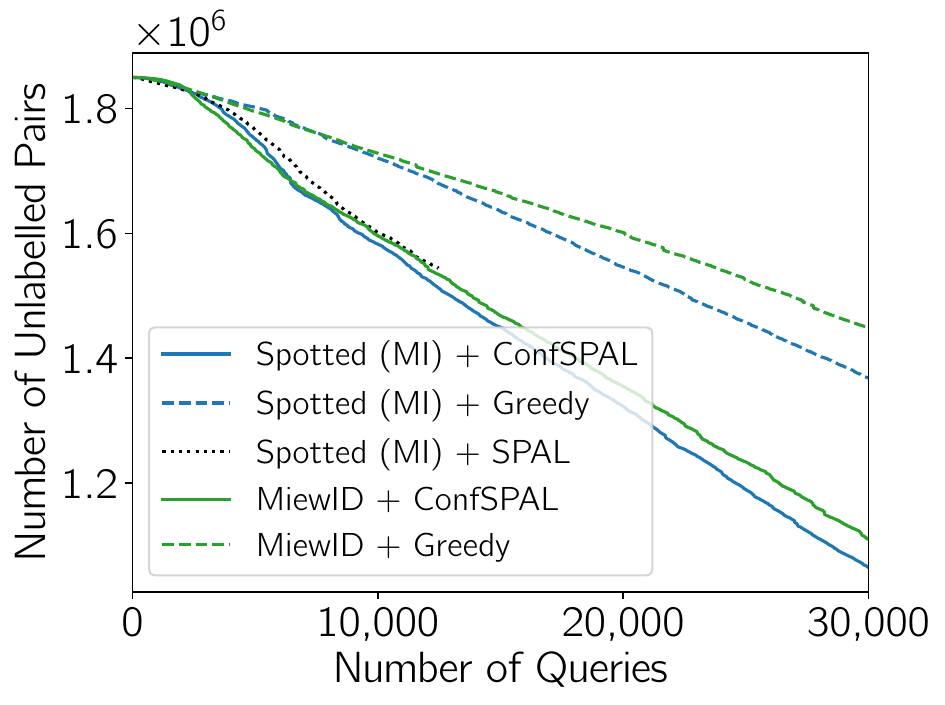}
        \caption{SpottedHyenaID415}
    \end{subfigure}
    
    \caption{Number of remaining unlabelled image pairs as a function of annotation queries for our three datasets. (a-c) MegaDescriptor (MD) is used as the frozen image backbone. (d-f) MiewID (MI) is used as the frozen image backbone. Faster reduction indicates improved annotation efficiency.}
    \label{fig:queryvunlabelled}
}
\end{figure}

\Cref{fig:queryvmatches,fig:queryvunlabelled} compare the cumulative number of positive matches discovered and the reduction in unlabeled pairs as a function of annotation queries. Our model \net~consistently outperforms both the MegaDescriptor and MiewID baselines, demonstrating the importance of spatio-temporal priors alongside visual similarity. \net{} (MI) + ConfSPAL reduces the number of human queries required to fully label the dataset by $69.2\%$ on LeopardID102 and $42.0\%$ on SpottedHyenaID109 compared to MiewID with greedy sampling, corresponding to a reduction from 31.8 to 9.8  and 37.4 to 21.7 labeling  hours for the LeopardID102 and SpottedHyenaID109 datasets respectively, extrapolating a measured annotation rate of 5.78 queries per minute. For the larger SpottedHyenaID415 datasets \net{} (MI) + ConfSPAL discovers $4070$ more positive matches than MiewID + Greedy within the same labeling budget of 86 hours, corresponding to $95.6\%$ of all positive matches discovered compared to only $52.4\%$ for MiewID + Greedy. Notably, even when the top-5 accuracy improvement on SpottedHyenaID109 is modest at 2pp, this translates to a substantial 42\% reduction in required human queries in a simulated labeling scenario, highlighting that small improvements in ranking quality can lead to  considerable gains in annotation efficiency. The impact of active sampling is especially noticeable for SpottedHyenaID415, the most challenging of our three datasets. Combining \net{} with our active sampling method yields $799$ and $500$ additional positive matches over \net{} with greedy sampling for the MegaDescriptor and MiewID backbones, respectively. The results also demonstrate the role of active sampling in the Re-ID pipeline. While our SPAL-based active sampling improves performance when combined with our fused embeddings, it alone is insufficient to compensate for weak underlying representations. In particular, while applying ConfSPAL to the MegaDescriptor or MiewID embeddings does consistently boost performance over greedy sampling, \net{} with greedy sampling still outperforms the baseline methods with ConfSPAL. This behavior can be attributed to the quality of the induced similarity structure. When the embedding space does not form sufficiently coherent clusters, boundary-based sampling strategies select pairs that lie near poorly defined cluster margins. Although these samples satisfy the heuristic criteria of being informative (e.g.\ dissimilar intra-cluster or similar inter-cluster pairs), they are not discriminative due to the weak cluster structure. This leads to inefficient querying and slower discovery of true matches. In contrast, the stronger separation induced by the adapted and fused embeddings yields more coherent clusters in the early stages of annotation. This allows our ConfSPAL to identify informative samples for labeling, thereby improving the tuning of the ConfSPAL MLP head. Across all three datasets, \net~reduces the total number of queries required to reach an equivalent (and in the case of the SpottedHyenaID415 dataset, greater) number of confirmed matches. Additionally, the rate at which matches are discovered is consistently higher, indicating improved annotation efficiency. We additionally compare against the vanilla SPAL sampling scheme, which omits the confidence-based sampling stage. As shown in~\cref{fig:queryvmatches,fig:queryvunlabelled}, vanilla SPAL tracks closely to ConfSPAL, but fails to propose new pairs once the cluster boundaries stabilize. This causes vanilla SPAL to terminate pair discovery before all matches have been found, whereas ConfSPAL's reversion to confidence-based sampling continues to propose new pairs beyond this point.

\section*{Conclusion}
We introduced \net, a location-informed, human-in-the-loop animal re-identification framework. Our method utilizes the structure of camera-trap deployments by combining image-based visual embeddings with a spatio-temporal feasibility score derived from camera geometry and timestamps. By using speed-based, continuous feasibility scores as matching cues and pseudo-labels for a lightweight head, our approach suppresses physically unlikely matches without requiring identity labels. Additionally, coupling our location-informed visual network with our active pair sampling strategy, reduces the number of required expert reviews to discover positive matches.
\net~ consistently improves top-$k$ retrieval performance over strong visual baselines across three challenging camera-trap datasets that we publish with this work. We encourage future work to release camera-trap locations and timestamps alongside public datasets, and to explore additional auxiliary cues as well as transductive learning and uncertainty to further boost animal reID performance.


\section*{Data Availability}
The three animal re-identification datasets generated and/or analysed during the current study are available in the Hugging Face repositories at https://doi.org/10.57967/hf/9288, https://doi.org/10.57967/hf/9290, and https://doi.org/10.57967/hf/9289.
The camera locations are released in relative coordinates to protect the privacy of the survey sites.

\bibliography{main.bib}

\section*{Acknowledgements}
All figures were prepared by H.S.K. with assistance from J.H.

\section*{Author Contributions}
H.S.K. led methodology and software development, visualization, and wrote the original draft. J.H. contributed to the technical discussion and assisted with the figures. K.H., L.H., B.M., K.N., J.S.S. and A.S. contributed to data curation. A.L. led funding acquisition and data curation. A.V. provided institutional support. M.W. led conceptualization, funding acquisition, and data curation.  D.D.M. contributed to conceptualization, technical discussion, and provided supervision. All authors contributed to writing-review and editing.

\section*{Funding}
H.S.K., A.L., J.S.S, A.S, and M.W. were funded by Allen Family Philanthropies.
J.H. was funded by the Deutsche Forschungsgemeinschaft (DFG, German Research Foundation) – SFB 1597 – 499552394 and J.H. acknowledges travel support from ELISE (GA no. 951847) and ELSA - European Lighthouse on Secure and Safe AI funded by the European Union under GA no. 101070617. We are grateful to Allen Family Philanthropies for funding this work.

\section*{Competing Interests}
The authors declare no competing interests.

\end{document}